\documentclass[11pt]{article}

\PassOptionsToPackage{table}{xcolor}
\usepackage[final]{acl}

\usepackage{times}
\usepackage{latexsym}
\usepackage[T1]{fontenc}
\usepackage[utf8]{inputenc}
\usepackage{microtype}
\usepackage{inconsolata}
\usepackage{booktabs}
\usepackage{enumitem}
\usepackage{amsmath}
\usepackage{multirow}
\usepackage{makecell}
\usepackage{array}
\usepackage{pifont}
\usepackage{graphicx}
\usepackage{textcomp}
\usepackage[ruled]{algorithm2e}
\usepackage{tcolorbox}
\tcbuselibrary{skins, breakable, listings}
\lstdefinestyle{promptlisting}{
    basicstyle=\ttfamily\scriptsize,
    breaklines=true,
    breakatwhitespace=false,
    columns=fullflexible,
    keepspaces=true,
    showstringspaces=false
}
\usepackage{tikz}
\usepackage{etoolbox}
\usepackage{fontawesome5}

\definecolor{uiGray100}{HTML}{F3F4F6}
\definecolor{uiGray200}{HTML}{E5E7EB}
\definecolor{uiGray400}{HTML}{9CA3AF}
\definecolor{uiGray500}{HTML}{6B7280}
\definecolor{uiGray800}{HTML}{1F2937}

\definecolor{uiBlue50}{HTML}{EFF6FF}
\definecolor{uiBlue100}{HTML}{DBEAFE}
\definecolor{uiBlue600}{HTML}{2563EB}
\definecolor{uiBlue700}{HTML}{1D4ED8}
\definecolor{uiBlue900}{HTML}{1E3A8A}

\definecolor{uiGreen50}{HTML}{F0FDF4}
\definecolor{uiGreen100}{HTML}{DCFCE7}
\definecolor{uiGreen500}{HTML}{22C55E}
\definecolor{uiGreen600}{HTML}{16A34A}
\definecolor{uiGreen700}{HTML}{15803D}

\definecolor{uiYellow50}{HTML}{FEFCE8}
\definecolor{uiYellow200}{HTML}{FEF08A}

\definecolor{uiRed50}{HTML}{FEF2F2}
\definecolor{uiRed100}{HTML}{FEE2E2}
\definecolor{uiRed500}{HTML}{EF4444}
\definecolor{uiRed700}{HTML}{B91C1C}

\definecolor{uiOrange100}{HTML}{FFEDD5}
\definecolor{uiOrange500}{HTML}{F97316}
\definecolor{uiOrange700}{HTML}{C2410C}

\definecolor{uiPurple100}{HTML}{F3E8FF}
\definecolor{uiPurple500}{HTML}{A855F7}
\definecolor{uiPurple700}{HTML}{7E22CE}

\definecolor{uiOrange50}{HTML}{FFF7ED}
\definecolor{uiOrange200}{HTML}{FED7AA}
\definecolor{uiOrange600}{HTML}{EA580C}
\colorlet{uiOrange}{uiOrange600}

\definecolor{uiAcadBlue}{HTML}{6A8CAF}   
\definecolor{uiAcadGreen}{HTML}{8FAD88}  
\definecolor{uiAcadPurple}{HTML}{A390B2} 
\definecolor{uiAcadTeal}{HTML}{72A0A5}   
\definecolor{uiAcadRose}{HTML}{C6878F}   
\definecolor{uiAcadSand}{HTML}{C2A878}

\colorlet{uiBlue}{uiBlue600}
\colorlet{uiGray}{uiGray500}
\colorlet{uiGreen}{uiGreen500}
\colorlet{uiYellow}{uiYellow200}
\colorlet{uiPurple}{uiPurple500}
\colorlet{uiRed}{uiRed500}
\colorlet{uiOrange}{uiOrange500}

\newtcolorbox{infobox}[2][]{
    enhanced, breakable, boxrule=0.5pt, arc=1.5mm,
    colback=#2!15, colframe=#2!30, coltext=uiGray800,
    boxsep=1mm, left=2mm, right=2mm, top=1mm, bottom=1mm, #1
}

\newcommand{\badge}[2]{%
    \tcbox[on line, boxsep=0pt, left=4pt, right=4pt, top=2pt, bottom=2pt, 
           colback=#1!20, colframe=#1!20, arc=1mm, boxrule=0pt]{\scriptsize \textcolor{#1!80!black}{\textbf{#2}}}
}

\newcommand{\titem}[4]{%
    \par\noindent
    \llap{\tikz[baseline=-0.5ex]\draw[#1, fill=white, line width=1.5pt] (0,0) circle (3pt);\hspace{14pt}}%
    {\small #2 \hfill \scriptsize \textcolor{uiGray500}{#3}}\par
    \ifstrempty{#4}{}{\vspace{1mm}#4\par}
    \vspace{3.5mm}
}

\newtcolorbox{warningbox}[1][]{
    enhanced, boxrule=0.5pt, arc=1mm,
    colback=uiOrange100!50, colframe=uiOrange500!50, coltext=uiOrange700,
    fontupper=\scriptsize\ttfamily,
    boxsep=0.5mm, left=2mm, right=2mm, top=1mm, bottom=1mm,
    before upper=\faExclamationTriangle\ , #1
}

\title{EviDx: Evidence-Aware Active Diagnosis with Scaffolded LLM Agents}
\author{
  \textbf{Lihang Zeng}$^{1}$,
  \textbf{Shaoting Zhang}$^{1*}$,
  \textbf{Xiaofan Zhang}$^{1,2}$\thanks{Corresponding authors}
\\
  $^{1}$Shanghai Jiao Tong University, $^{2}$Shanghai Innovation Institute
}

\begin{document}

\maketitle
\begin{abstract}
Clinical diagnosis is an active evidence-seeking process in which clinicians acquire evidence, update competing hypotheses, and decide when the available evidence is sufficient for diagnosis. Yet many medical diagnosis systems built around large language models (LLMs) still formulate diagnosis as static case-to-answer prediction, with limited support for evidence acquisition. Agentic LLMs offer a dynamic alternative through tool use and intermediate diagnostic trajectories, but existing systems often under-specify how patient evidence should be exposed, scaffolded, and controlled at runtime. We introduce EviDx, an evidence-aware active diagnosis framework that pairs patient-specific diagnostic environments with a clinical diagnostic scaffold and an observer-guided runtime harness. In EviDx, $\mathcal{E}$-Synthesis constructs interactive environments from raw clinical cases; the scaffold organizes role-specialized agents, evidence tools, and evolving evidence states; and the harness regulates diagnostic termination by tracking uncertainty and evidence coverage. A 3-level evaluation pyramid assesses execution robustness, reasoning dynamics, and diagnostic outcomes. Experiments show that EviDx improves diagnostic performance and process stability while revealing model-dependent capability boundaries.
\end{abstract}

\newcommand{\cmark}{\ding{51}}  
\newcommand{\xmark}{\ding{55}} 

\begin{table*}[t]
  \centering
  \resizebox{\textwidth}{!}{%
    \begin{tabular}{lccccc}
    \toprule
    \multirow{2}{*}{Work} & \multicolumn{2}{c}{Diagnosis Setting} & \multicolumn{2}{c}{System Design} & Evaluation \\
    \cmidrule(lr){2-3} \cmidrule(lr){4-5} \cmidrule(lr){6-6}
    & Interactive Env. & Evidence Access & Clinical Scaffold & Runtime Harness & Process-Level Eval. \\
    \midrule
    \textbf{MDAgents \citep{kim2024mdagents}} & \xmark & \cmark & \cmark & \xmark & \xmark \\
    \textbf{Agent Hospital \citep{li2024agent}} & \cmark & \xmark & \cmark  & \xmark & \xmark  \\
    \textbf{AgentClinic \citep{schmidgall2024agentclinic}} & \cmark & \cmark & \cmark  & \xmark & \cmark  \\
    \textbf{MedAgentBench \citep{jiang2025medagentbench}} & \cmark & \cmark & \xmark & \xmark & \cmark \\
    \textbf{AutoMedic \citep{oh2025automedic}} & \cmark & \xmark & \cmark & \xmark & \cmark \\
    \midrule
    \rowcolor{gray!10} \textbf{EviDx} & \textbf{\cmark} & \textbf{\cmark} & \textbf{\cmark} & \textbf{\cmark} & \textbf{\cmark} \\
    \bottomrule
    \end{tabular}%
  }
  \caption{Qualitative comparison of representative clinical LLM diagnosis systems and medical-agent evaluation environments. We summarize each work along design axes relevant to evidence-aware active diagnosis, including interactive evidence acquisition, clinical scaffolding, runtime control, and process-level evaluation.}
  \label{tab:medical_agents}
\end{table*}

\section{Introduction}

Clinical diagnosis (Dx) is not a single-step prediction problem. It is an active evidence-seeking process in which clinicians iteratively acquire evidence, integrate new findings into a differential diagnosis, refine competing hypotheses, and decide whether the current evidence is sufficient for diagnosis or whether further information should be gathered \citep{sackett1997evidence, pauker1980threshold}. A diagnostic conclusion is therefore meaningful only when it is supported by an adequate evidence trajectory, not merely by a plausible final answer.

Many existing medical diagnosis systems built around LLMs still formulate diagnosis as static case-to-answer prediction. A model receives a complete clinical vignette or answer set upfront, and then directly produces a diagnosis or selects from multiple choices. Such methods can elicit medical knowledge and final diagnoses, but they collapse evidence acquisition, evidence integration, and termination into a single prompt. As a result, they under-specify process failures that arise in active diagnosis, such as premature commitment before sufficient evidence has been acquired.

Tool-augmented and agentic medical systems offer a path toward dynamic diagnosis because they can query patient records and retrieve external medical knowledge \citep{schmidgall2024agentclinic, jiang2025medagentbench, wang2025medagent}. However, dynamic diagnosis is not obtained by tool access alone. The capability of an LLM agent is shaped by the environment that exposes patient evidence, the scaffold that organizes evidence actions, clinical roles, and evidence states, and the harness that monitors execution at runtime. Existing medical-agent systems often organize interaction through fixed workflows, turn budgets or consensus checks \citep{oh2025automedic, chen2025enhancing}, leaving evidence sufficiency and termination control only weakly modeled. This raises a general question: \textit{Can clinical LLM agents become more capable when interactive diagnostic environments are coupled with explicit evidence scaffolds and runtime harnesses for active diagnosis?}

We introduce \textbf{EviDx}, an evidence-aware active diagnosis framework for scaffolded and harnessed clinical LLM agents. First, $\mathcal{E}$-Synthesis converts raw clinical text into a patient-specific interactive environment in which evidence can be acquired through standardized interfaces. Second, the Clinical Dx Scaffold organizes diagnostic execution through role-specialized agents, evidence tools, and evolving evidence states, including patient-record and external-knowledge access through Model Context Protocol (MCP; \citealp{Anthropic2024ModelContext}) tool interfaces. Third, the Observer-Guided Harness monitors diagnostic uncertainty and runtime evidence coverage from the evolving diagnostic states to regulate diagnostic termination decisions. Finally, the 3-level evaluation pyramid evaluates the resulting trajectories across execution robustness, reasoning dynamics, and diagnostic outcomes.

Our main contributions are as follows:

\begin{itemize}[leftmargin=14pt]
\item We propose EviDx, an evidence-aware active diagnosis framework that transforms static clinical cases into interactive diagnostic environments and scaffolds LLM agents with clinical roles, evidence tools, and evidence states.
\item We introduce an Observer-Guided Harness that controls diagnostic progression by tracking uncertainty and evidence coverage, enabling explicit diagnostic termination decisions.
\item We design a 3-level evaluation pyramid to assess EviDx across execution robustness, reasoning dynamics, and outcomes. Experiments show that EviDx improves diagnostic performance and process stability while revealing model-dependent capability boundaries. Code and resources are available.\footnote{\url{https://github.com/LihangZeng/EviDx}}
\end{itemize}

\section{Related Work}

\noindent\textbf{LLMs for Clinical Diagnosis.}
LLMs have demonstrated strong performance on medical question answering and diagnosis-oriented benchmarks, including real-world clinical cases and expert-level multiple-choice settings \citep{chen2025benchmarking, zuo2025medxpertqa, zhu2025diagnosisarena}. These settings can elicit medical knowledge and final-answer behavior, but they often expose the case as a fixed input and ask the model to generate the diagnosis in one pass. In contrast, clinical diagnosis is an evidence-seeking process: physicians iteratively acquire patient information, integrate findings into a differential diagnosis, and decide whether the current evidence is sufficient. This motivates diagnosis settings that represent clinical cases as evolving evidence contexts.

\noindent\textbf{Tool-Augmented and Agentic LLMs.}
Tool-augmented medical agents seek to move beyond static QA by allowing models to retrieve patient data, consult external medical knowledge, or coordinate role-specialized reasoning \citep{wang2025medagent, li2024agent, wu2025kamac}. Interactive clinical environments and EHR-oriented benchmarks further stress that medical agents must act within patient-specific contexts rather than only answer fixed prompts \citep{schmidgall2024agentclinic, jiang2025medagentbench, liu2026physicianbench}. These works motivate the shift from answer generation to evidence seeking, while also exposing a remaining control problem: tool access must be coupled with decisions about which evidence is still missing and when the diagnostic process should terminate.

\noindent\textbf{Scaffolds and Harnesses for LLM Agents.}
Recent LLM-agent research increasingly treats agents as executable systems embedded in environments, where planning, tool use, memory or state, and trace-level evaluation are core design concerns \citep{yehudai2025survey, wei2026agentic,chen2025toolforge}. At the same time, runtime-enforcement work frames agent reliability as an execution-control problem: external monitors can inspect traces, apply constraints, and intervene before unsafe or invalid actions are committed \citep{wang2025agentspec, wang2025probguard}. In this literature, scaffolds provide structured support around an agent, including roles, tools, state representations, and workflows, while harnesses provide runtime machinery for monitoring and controlling execution. This distinction is useful for active diagnosis because evidence acquisition and termination are coupled but separable: one design layer organizes how an agent seeks and records evidence, while another decides whether the evolving trajectory is sufficient for termination.

\noindent\textbf{Evaluation of Clinical Reasoning and Medical Agents.}
Outcome-only evaluation can hide whether a model acquired the right evidence, used it correctly, or merely guessed the final answer. Recent medical-agent benchmarks therefore move toward interactive EHR tasks, virtual patient conversations, and execution-grounded evaluation \citep{oh2025automedic, jiang2025medagentbench, liu2026physicianbench}. For active diagnosis, evaluation should therefore distinguish execution robustness, reasoning dynamics, and diagnostic outcomes instead of collapsing the entire process into final accuracy.

Table~\ref{tab:medical_agents} makes these distinctions explicit by comparing representative systems along the axes emphasized above: interactive evidence acquisition, clinical scaffolding, runtime control, and process-level evaluation.

\section{Problem Formulation}
\label{sec:problem_formulation}

We formulate clinical LLM diagnosis as active differential diagnosis: an iterative process in which a model acquires evidence, updates a differential diagnosis, and decides when the accumulated trajectory is sufficient for a final answer.

\vspace{1mm}
\noindent\textbf{Active Differential Diagnosis.}
Let $x$ denote a clinical case and let $\mathcal{E}_x$ denote an interactive diagnostic environment associated with the patient. At step $t$, the system takes an evidence-seeking action $a_t$, receives an observation $r_t$ from $\mathcal{E}_x$, and appends it to the diagnostic trajectory $\mathcal{T}_t=\{x_0,(a_1,r_1),\ldots,(a_t,r_t)\}$. The system maintains a differential diagnosis set $D_t \subseteq D$ and reports a normalized diagnostic belief:
\vspace{-3pt}
{
\setlength{\belowdisplayskip}{2pt}
\begin{equation}
\resizebox{0.89\linewidth}{!}{$\displaystyle
\widehat{P}_t =
\{\hat{p}_t(d_i \mid \mathcal{T}_t)\}_{d_i \in D_t},
\sum_{d_i \in D_t}\hat{p}_t(d_i \mid \mathcal{T}_t)=1.
$}
\end{equation}
}
Here, $\widehat{P}_t$ represents the model-reported belief over competing diagnostic hypotheses. Active diagnosis therefore consists of evidence acquisition, evidence integration, diagnosis refinement, and termination decisions over the evolving trajectory $\mathcal{T}_t$.

\vspace{1mm}
\noindent\textbf{Evidence Sufficiency and Termination Decisions.}
At each step, the central decision is whether the current trajectory is sufficient for diagnosis or whether additional evidence should be acquired. We characterize diagnostic uncertainty using the entropy of the reported differential diagnosis:
\vspace{-3pt}
{
\setlength{\belowdisplayskip}{2pt}
\begin{equation}
H(t)=
-\sum_{d_i \in D_t}
\hat{p}_t(d_i \mid \mathcal{T}_t)
\log_2 \hat{p}_t(d_i \mid \mathcal{T}_t).
\label{eq:entropy}
\end{equation}
}
A lower $H(t)$ indicates a more concentrated belief state, while a higher $H(t)$ suggests that the differential remains unresolved. However, diagnostic confidence and evidence sufficiency are distinct: a model may report a confident belief before acquiring enough patient-specific evidence. The active diagnosis problem therefore requires termination decisions that consider both diagnostic resolution and the evidential support accumulated in the trajectory.

\section{Proposed Methodology}

EviDx contains four components: $\mathcal{E}$-Synthesis, Clinical Dx Scaffold, Observer-Guided Harness, and a 3-level evaluation pyramid. Together, they convert static clinical cases into active diagnostic environments, structure evidence acquisition, regulate termination, and evaluate the resulting diagnostic trajectories.

\begin{figure*}[t]
  \centering
  \includegraphics[width=\textwidth]{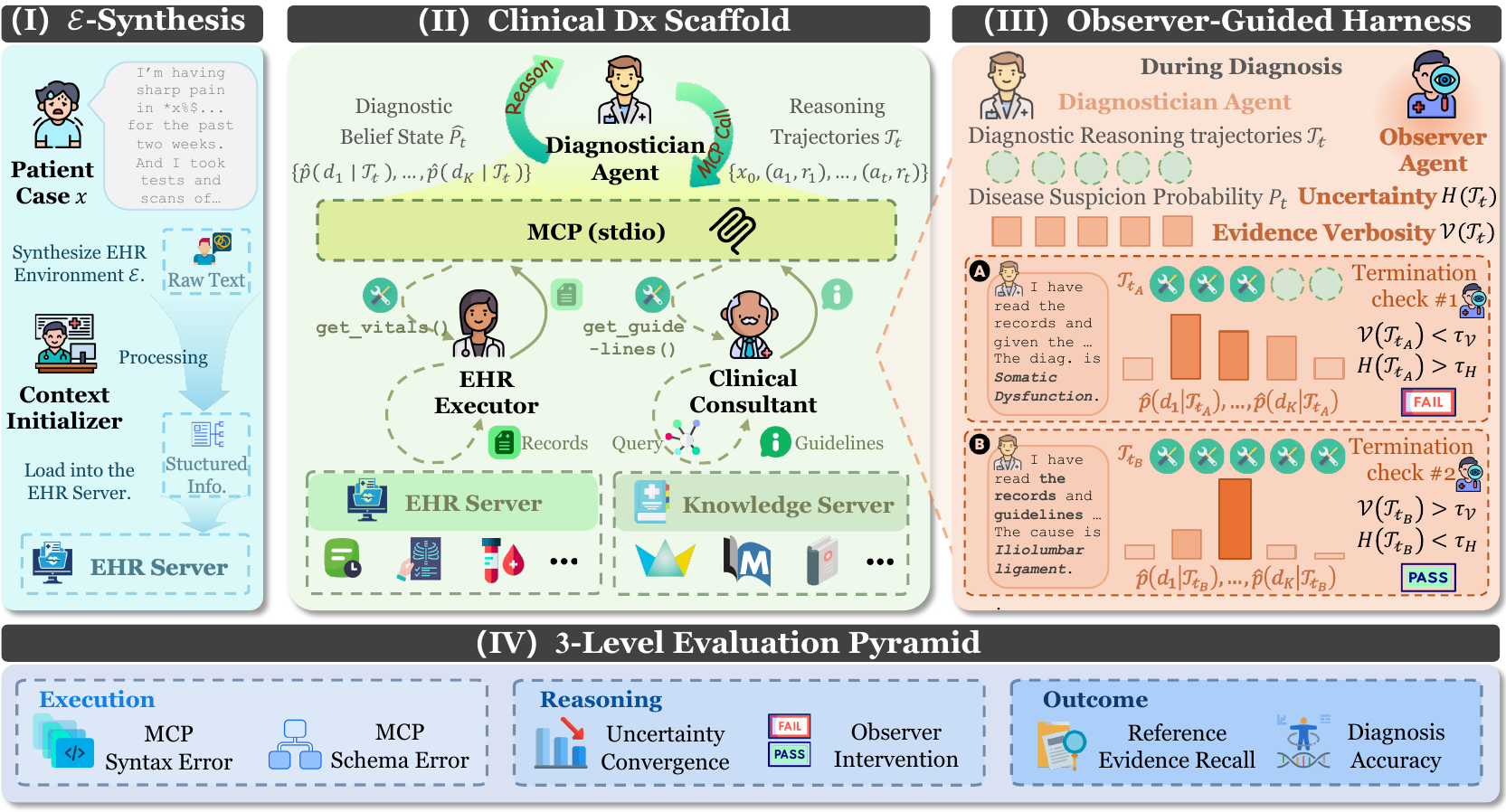}
  \caption{Overview of EviDx. \textbf{(I)} $\mathcal{E}$-Synthesis converts raw clinical text into a patient-specific diagnostic environment through the Context Initializer. \textbf{(II)} The Clinical Dx Scaffold structures evidence acquisition and diagnostic reasoning through clinical roles, evidence tools, and evidence states. \textbf{(III)} The Observer-Guided Harness monitors uncertainty $H(t)$ and runtime evidence coverage $\mathcal{V}(t)$ to govern diagnostic termination. \textbf{(IV)} The 3-level evaluation pyramid evaluates execution robustness, reasoning dynamics, and diagnostic outcomes.}
  \label{fig:overview}
\end{figure*}

\subsection{\texorpdfstring{$\mathcal{E}$-Synthesis}{E-Synthesis}}
\label{subsec:e_synthesis}

EviDx begins by converting a static clinical case into a patient-specific diagnostic environment. For each patient case $x$, the Context Initializer processes the raw clinical text, converts it into structured patient data, and loads these records into the EHR Server. $\mathcal{E}_x$ then couples this patient-record backend with evidence-access tools, forming an interactive environment in which subsequent diagnostic actions can query patient-specific records and external medical knowledge.

The Context Initializer preserves clinically relevant ambiguity, qualifiers and negative findings, so that structuring the case does not collapse the evidence available for diagnosis. In this way, $\mathcal{E}$-Synthesis changes the diagnosis setting from static text consumption to active evidence acquisition.

\subsection{Clinical Dx Scaffold}
\label{subsec:dx_scaffold}

The Clinical Dx Scaffold structures how the LLM interacts with $\mathcal{E}_x$. It consists of role-specialized components, evidence states, and an active diagnostic workflow rather than generic role prompting. The Diagnostician maintains the differential diagnosis $D_t$, the reported belief state $\widehat{P}_t$, and the trajectory $\mathcal{T}_t$. At each step, it identifies the next information need and issues an evidence action as a Model Context Protocol (MCP; \citealp{Anthropic2024ModelContext}) tool call, which standardizes tool invocation and data exchange. The EHR Executor and Clinical Consultant handle these calls by retrieving patient records or external evidence, respectively, and return an observation $r_t$ to the Diagnostician.

\noindent\textbf{Diagnostician.}
The Diagnostician is the central reasoning component. Given $\mathcal{T}_{t-1}$, it updates the differential diagnosis, reports $\widehat{P}_t$, and decides whether to query patient data, consult medical knowledge, or attempt to finish.

\noindent\textbf{EHR Executor.}
The EHR Executor handles deterministic access to the EHR Server within $\mathcal{E}_x$, retrieving requested records such as demographics, history, labs, and imaging reports, and returning them as standardized observations.

\noindent\textbf{Clinical Consultant.}
The Clinical Consultant handles external evidence access. For knowledge-directed actions, it decomposes the request, retrieves relevant literature from the Knowledge Server using hybrid dense and BM25 sparse retrieval following MedRAG \citep{xiong2024benchmarking}, and returns a response to the Diagnostician.

\subsection{Observer-Guided Harness}
\label{subsec:observer}

A scaffold alone is insufficient for active diagnosis: even when agents can request evidence, the system still needs runtime control over whether the current trajectory is ready to terminate. The Observer-Guided Harness provides this control. Whenever the Diagnostician proposes to finish, the harness evaluates whether the trajectory is both sufficiently resolved and sufficiently explored according to runtime evidence coverage. If not, it returns feedback and forces continued evidence acquisition.

\vspace{1mm}
\noindent\textbf{Diagnostic Uncertainty.}
The harness uses $H(t)$ from Eq.~\eqref{eq:entropy} to measure whether the differential diagnosis remains unresolved. High entropy indicates that probability mass is still distributed across competing diagnoses, while low entropy indicates that the belief state has concentrated.

\vspace{1mm}
\noindent\textbf{Runtime Evidence Coverage.}
An agent may report confidence before exploring enough patient-specific information, so the harness also tracks runtime evidence coverage $\mathcal{V}(\mathcal{T}_t)$, which measures the fraction of available EHR sections explicitly queried. Let $S_{\text{Available}}$ be the set of EHR sections exposed by the runtime EHR menu, and let $S_{\text{Queried}}(\mathcal{T}_t)$ be the subset accessed through tool calls up to step $t$:
\vspace{-3pt}
{
\setlength{\belowdisplayskip}{2pt}
\begin{equation}
\mathcal{V}(\mathcal{T}_t)=
\frac{|S_{\text{Queried}}(\mathcal{T}_t) \cap S_{\text{Available}}|}
{|S_{\text{Available}}|}.
\end{equation}
}
Unlike post-hoc evidence metrics used for evaluation, $\mathcal{V}$ is computed prospectively from runtime-accessible metadata and tool-call traces.

\vspace{1mm}
\noindent\textbf{Termination Criteria.}
The harness approves termination only when the trajectory satisfies both uncertainty and evidence coverage criteria:
\vspace{-5pt}
\begin{equation}
\mathcal{B}(\mathcal{T}_t)=
\big[H(t)<\tau_H(t)\big]\land
\big[\mathcal{V}(\mathcal{T}_t)>\tau_\mathcal{V}(t)\big].
\end{equation}
Here, $\mathcal{B}(\mathcal{T}_t)=\text{True}$ means termination is approved; otherwise the system continues. The thresholds $\tau_H(t)$ and $\tau_\mathcal{V}(t)$ specify step-aware criteria for approving termination: as the interaction proceeds, the harness gradually relaxes the entropy requirement and, after sufficient search, lowers the minimum coverage required for approval. The exact schedules are given in Appendix~\ref{app:implementation}. A hard budget $T_{\max}$ provides a final safeguard against loops.

\begin{algorithm}[t]
    \small
    \caption{EviDx Active Diagnosis Loop}
    \label{alg:evidx_loop}
    \KwIn{Raw patient case $x$}
    \KwOut{Final diagnosis report}
    $(x_0,\mathcal{E}_x) \leftarrow \textsc{$\mathcal{E}$-Synthesis}(x)$ \\
    $\mathcal{T}_0 \leftarrow [x_0]$ \\
    \For{$t = 1$ \KwTo $T_{\max}$}{
        $(D_t, \widehat{P}_t, a_t) \leftarrow \textsc{Diagnostician}(\mathcal{T}_{t-1})$ \\
        \uIf{$a_t = \texttt{finish}$}{
            $b_t \leftarrow \textsc{ObserverHarness}(\mathcal{T}_{t-1}, \widehat{P}_t)$ \\
            \If{$b_t = \texttt{terminate}$}{
                \Return final diagnosis report
            }
            $r_t \leftarrow b_t$ 
        }
        \uElseIf{$a_t$ targets patient data}{
            $r_t \leftarrow \textsc{EHRExecutor}(a_t)$
        }
        \Else{
            $r_t \leftarrow \textsc{ClinicalConsultant}(a_t)$
        }
        $\mathcal{T}_t \leftarrow \textsc{Append}(\mathcal{T}_{t-1}, a_t, r_t)$
    }
    \Return final diagnosis
\end{algorithm}

Algorithm~\ref{alg:evidx_loop} summarizes the inference process. EviDx first synthesizes $\mathcal{E}_x$ from the raw case, initializes the trajectory, and then iterates between Diagnostician actions, patient-data retrieval, medical-knowledge consultation, and harness-controlled termination decisions.

\subsection{3-Level Evaluation Pyramid}
\label{subsec:eval}

Evaluating active diagnostic agents solely on final accuracy obscures the underlying mechanics of evidence acquisition, reasoning, and tool use. EviDx therefore uses a 3-level evaluation pyramid to assess diagnostic trajectories across execution robustness, reasoning dynamics, and outcomes.

\vspace{1mm}
\noindent\textbf{Level 1: Execution Robustness.} \\
This level evaluates whether the model can correctly interact with the tool interface. It uses the following execution metrics:
\begin{itemize}[leftmargin=*, topsep=0pt, itemsep=2pt, parsep=0pt]
\item \textit{MCP Syntax Error:} Measures formatting failures during tool invocation, such as malformed JSON payloads, missing brackets, or hallucinated parameter names.
\item \textit{MCP Schema Error:} Captures cases where the model fails to use the native tool-calling schema, instead outputting raw natural language when an MCP action was required.
\end{itemize}

\vspace{1mm}
\noindent\textbf{Level 2: Reasoning Dynamics.} \\
This level examines how the agent manages the diagnostic process over time, including belief concentration and harness intervention.
\begin{itemize}[leftmargin=*, topsep=0pt, itemsep=2pt, parsep=0pt]
    \item \textit{Uncertainty Convergence (Unc.)} Measures how effectively the agent assimilates information over time. We compute the normalized average of step-wise entropy:
    \vspace{-5pt}
    {
    \setlength{\belowdisplayskip}{2pt}
    \begin{equation}
    \text{Unc.} = \frac{1}{T} \sum_{t=1}^{T} \frac{H(t)}{H_{init}}
    \end{equation}
    }
    Here, $H_{init}$ denotes the entropy reported after the Diagnostician receives the initial patient information, before subsequent evidence acquisition. A lower score indicates faster concentration of the model-reported belief state, but does not by itself imply correctness or calibration.

    \item \textit{Observer Intervention (Obs.):} Records the frequency of intercepts triggered by the Observer. A high intervention count indicates more frequent attempted early termination under the current harness criteria. We interpret it as a descriptive process-control signal rather than a standalone measure of reasoning quality.
\end{itemize}

\vspace{1mm}
\noindent\textbf{Level 3: Diagnostic Outcomes.} \\
This level evaluates whether the agent produces the correct diagnosis and whether the final reasoning is grounded in the right evidence.

\begin{itemize}[leftmargin=*, topsep=0pt, itemsep=2pt, parsep=0pt]
\item \textit{Reference Evidence Recall (Acq. and Cog.).} We introduce a bipartite metric to distinguish between retrieving evidence and reasoning over it.
(i) \textit{Acquisition} verifies whether the agent successfully retrieved predefined reference evidence into its trajectory $\mathcal{T}_t$ via MCP tools, using exact character matching and LLM-based semantic matching.
(ii) \textit{Cognition} evaluates whether that evidence was actively utilized in the final diagnostic reasoning. This distinguishes models that retrieve evidence from those that incorporate it into their reasoning, a failure mode traditional accuracy metrics would miss.

\item \textit{Diagnosis Accuracy (Open and MC).} Standard multiple-choice QA (MCQA) can inflate scores via heuristic cues. To avoid this, we adopt a progressive dual-track design: the agent first generates an open-ended diagnosis using the \texttt{finish} tool, graded by an LLM judge, and is then evaluated on a multiple-choice task. This ordering reduces the chance that MC performance is driven solely by option-elimination heuristics.
\end{itemize}

\section{Experimental Setup}
\label{sec:experimental_setup}

\subsection{Datasets \& Models}
\label{subsec:datasets}

To evaluate diagnostic capabilities across diverse clinical scenarios, we select three challenging medical diagnosis datasets: (1) JAMA \citep{chen2025benchmarking}, containing 1,524 real-world clinical cases from the JAMA Network with expert-written explanations; (2) MedXpertQA-Text \citep{zuo2025medxpertqa}, a text-only subset of 2,455 expert-level multiple-choice questions, from which we further curate MedXpertQA-Diag, a subset of 234 diagnostic-focused questions; and (3) DiagnosisArena \citep{zhu2025diagnosisarena}, comprising 1,113 structured clinical cases from top-tier medical journals for open-ended diagnostic reasoning. These datasets span real-world cases and expert-level diagnostic reasoning settings. Because EviDx requires multi-step tool interaction and trajectory-level evaluation for each case, we conduct a controlled evaluation on a fixed random subset of 100 instances from each dataset, using the same sampled cases across all methods and models.

We evaluate several large-scale language models: GPT-5.2 \citep{openai_gpt5_intro}, Claude Sonnet 4.6 \citep{anthropic_claude_sonnet}, GLM-5 \citep{zeng2026glm}, and DeepSeek-V3.2 \citep{liu2025deepseek}, as well as smaller models: Ministral3-8B \citep{liu2026ministral}, Qwen3-8B \citep{yang2025qwen3}, and Llama3.1-8B \citep{grattafiori2024llama}. For Clinical Consultant, we use Qwen3-embedding-4B \citep{zhang2025qwen3} for dense retrieval. Details are in Section~\ref{app:implementation}.

\subsection{Reference Evidence Curation}
\label{subsec:golden_evidence}

\begin{figure}[t]
    \centering
    \includegraphics[width=\columnwidth]{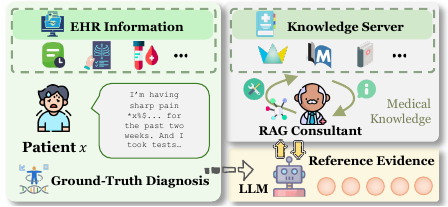}
    \caption{Reference evidence curation pipeline. The figure illustrates the automated extraction stage; after extraction, a subset of annotations is reviewed by physicians for clinical plausibility and evidence relevance, and the research team checks adherence and validity.}
    \label{fig:golden}
\end{figure}

To enable Reference Evidence Recall in Section~\ref{subsec:eval}, we construct an extractive curation pipeline that maps each case to its diagnosis-critical evidence. Given the raw patient text and the ground-truth diagnosis, GPT-5.2 first plans targeted diagnostic queries, then uses the Clinical Consultant to retrieve supporting knowledge, and finally extracts the bottleneck evidence set from the case. The extracted evidence is intended to capture the minimal clinically relevant clues required to support the ground-truth diagnosis and distinguish it from competing differentials.

Our pipeline is designed to be extractive rather than free-form generative. In the final annotation stage, the model is constrained to output exact substrings from the original case text, while grounding the extraction in retrieved medical references. This design reduces unsupported evidence synthesis and keeps the annotated clues directly verifiable against the source case. 
After automated extraction, we conduct a physician review on a subset of annotations. Physicians assess whether each extracted evidence span is present in the original case, clinically relevant to the reference diagnosis, and useful for distinguishing the diagnosis from plausible differentials. This review is intended as a plausibility check rather than full clinician adjudication; therefore, we treat Med-Evidence-2.6k as an LLM-assisted reference benchmark for evaluating evidence-seeking behavior, not as a clinician-adjudicated gold standard. Additional details of the review protocol are provided in Appendix~\ref{app:dataset_curation}.
\section{Results \& Analysis}

We first establish execution robustness (Level 1) in Section~\ref{exe}, then present diagnostic outcomes (Level 3) in Section~\ref{main}, followed by an analysis of the cognitive dynamics and observer harness (Level 2) that drive these outcomes in Section~\ref{cog}.

\subsection{Execution Robustness}
\label{exe}
\begin{figure*}[t]
  \centering
  \includegraphics[width=\textwidth]{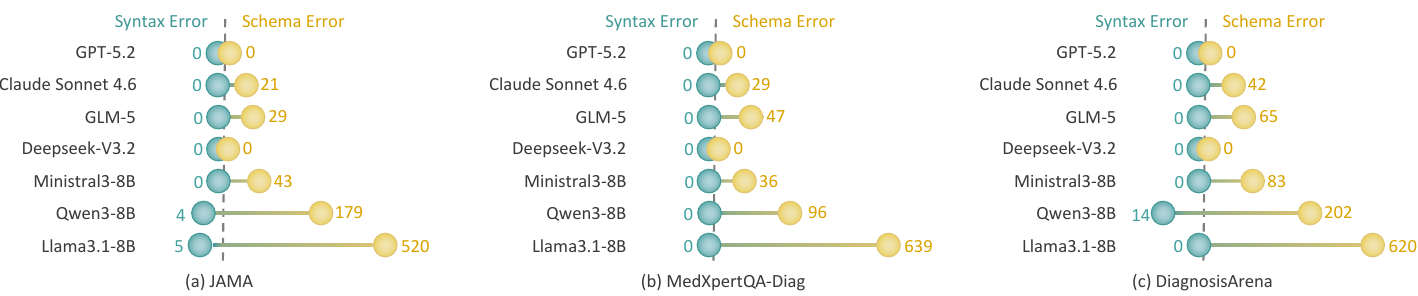}
  \caption{Execution Robustness (Level 1). Distribution of MCP Syntax and Schema errors.}
  \label{fig:level1}
\end{figure*}

Reliable information-gathering is the prerequisite for evidence-based reasoning. Figure~\ref{fig:level1} shows a clear capability gap in schema-constrained tool use. GPT-5.2 and DeepSeek-V3.2 complete the evaluated runs without syntax or schema failures, while Claude Sonnet 4.6 and GLM-5 occasionally fall back to natural language instead of native tool calling. The 8B-parameter open-weight models accumulate more schema failures, with Llama3.1-8B exceeding 500 failure events. This suggests that execution robustness is not merely a formatting issue, but a prerequisite for preserving the evidence-acquisition loop. Because such errors can interrupt evidence acquisition, Level 1 evaluation is necessary for diagnosing whether an interactive system can maintain the tool-use loop at all.

\subsection{Main Outcomes}
\label{main}

We evaluate diagnostic outcomes by examining final answers and supporting evidence. As shown in Table~\ref{tab:result}, EviDx improves MC diagnostic performance and process stability. These gains align with the central design of EviDx: patient-specific environments expose evidence, the Clinical Dx Scaffold routes patient-record and knowledge actions, and the Observer-Guided Harness constrains premature termination. The improvements are clearest in MC diagnosis, while open-ended diagnosis remains limited by base-model medical reasoning.

\noindent\textbf{EviDx Improves MC Diagnostic Performance.}
Following our Level 3 evaluation, the full pipeline achieves the highest MC accuracy among the evaluated settings on several datasets; for example, Claude Sonnet 4.6 reaches 66\% on JAMA and 52\% on DiagnosisArena. The improvement is most evident for 8B-parameter open-weight models: on JAMA, Qwen3-8B increases from 18\% to 44\%, while Llama3.1-8B increases from 17\% to 31\%. These gains suggest that scaffolded access to patient-specific and external evidence helps the Diagnostician before final prediction.

\begin{table*}[t]
\centering
\definecolor{lightgray}{gray}{0.9}
\resizebox{\textwidth}{!}{%
\begin{tabular}{l l c c c c c c c c c c c c c c c c c c}
\toprule
& & \multicolumn{6}{c}{JAMA} 
& \multicolumn{6}{c}{MedXpertQA-Diag} 
& \multicolumn{6}{c}{DiagnosisArena} \\
\cmidrule(lr){3-8} \cmidrule(lr){9-14} \cmidrule(lr){15-20}
& & \multicolumn{2}{c}{Diag. Acc} 
& \multicolumn{2}{c}{Evid. Rec.} 
& \multicolumn{2}{c}{Reas.} 
& \multicolumn{2}{c}{Diag. Acc} 
& \multicolumn{2}{c}{Evid. Rec.} 
& \multicolumn{2}{c}{Reas.} 
& \multicolumn{2}{c}{Diag. Acc} 
& \multicolumn{2}{c}{Evid. Rec.} 
& \multicolumn{2}{c}{Reas.} \\
\cmidrule(lr){3-4} \cmidrule(lr){5-6} \cmidrule(lr){7-8} 
\cmidrule(lr){9-10} \cmidrule(lr){11-12} \cmidrule(lr){13-14} 
\cmidrule(lr){15-16} \cmidrule(lr){17-18} \cmidrule(lr){19-20}
Models & Method
& Open$\uparrow$ & MC$\uparrow$ & Acq. & Cog. & Obs.$\downarrow$ & Unc. 
& Open$\uparrow$ & MC$\uparrow$ & Acq. & Cog. & Obs.$\downarrow$ & Unc. 
& Open$\uparrow$ & MC$\uparrow$ & Acq. & Cog. & Obs.$\downarrow$ & Unc. \\
\midrule
\rowcolor{lightgray} \multicolumn{20}{c}{\textit{\textbf{Large-Scale Models}}} \\
\multirow{3}{*}{GPT-5.2} 
& Single Agent   & 41 & 54 & 100 & 62.2 & -- & -- & 42 & 42 & 100 & 81.8 & -- & -- & 36 & 40 & 100 & 74.9 & -- & -- \\
& EviDx (wo Obs.) & 40 & 59 & 92.8 & 63.4 & -- & 0.94 & 42 & 41 & 94.0 & 71.9 & -- & 0.89 & 37 & 43 & 95.7 & 68.5 & -- & 0.84 \\
& EviDx           & 44 & 63 & 93.9 & 64.0 & 62 & 0.86 & 46 & 45 & 94.8 & 72.6 & 45 & 0.88 & 42 & 46 & 96.4 & 69.8 & 86 & 0.83 \\
\cline{1-20}
\multirow{3}{*}{Claude Sonnet 4.6} 
& Single Agent   & 42 & 53 & 100 & 70.4 & -- & -- & 32 & 36 & 100 & 88.0 & -- & -- & 37 & 51 & 99.9 & 75.5 & -- & -- \\
& EviDx (wo Obs.) & 46 & 62 & 95.5 & 68.7 & -- & 0.89 & 33 & 39 & 95.1 & 71.2 & -- & 0.86 & 41 & 46 & 94.9 & 67.5 & -- & 0.81 \\
& EviDx           & 50 & 66 & 96.7 & 70.9 & 63 & 0.87 & 38 & 40 & 97.8 & 73.3 & 51 & 0.83 & 45 & 52 & 96.1 & 69.2 & 64 & 0.79 \\
\cline{1-20}
\multirow{3}{*}{GLM-5} 
& Single Agent   & 22 & 45 & 100 & 62.4 & -- & -- & 30 & 35 & 100 & 76.7 & -- & -- & 26 & 29 & 100 & 75.0 & -- & -- \\
& EviDx (wo Obs.) & 21 & 57 & 85.8 & 64.5 & -- & 0.88 & 31 & 37 & 93.5 & 63.9 & -- & 0.91 & 28 & 41 & 92.1 & 67.8 & -- & 0.88 \\
& EviDx           & 25 & 57 & 87.1 & 64.8 & 67 & 0.87 & 37 & 38 & 94.2 & 65.0 & 94 & 0.89 & 34 & 42 & 92.8 & 68.4 & 39 & 0.86 \\
\cline{1-20}
\multirow{3}{*}{DeepSeek-V3.2} 
& Single Agent   & 39 & 52 & 100 & 58.7 & -- & -- & 24 & 25 & 99.6 & 70.4 & -- & -- & 24 & 32 & 99.1 & 58.6 & -- & -- \\
& EviDx (wo Obs.) & 40 & 56 & 89.8 & 59.8 & -- & 0.80 & 30 & 32 & 87.5 & 65.7 & -- & 0.81 & 24 & 29 & 86.4 & 61.7 & -- & 0.77 \\
& EviDx           & 43 & 58 & 91.8 & 61.6 & 15 & 0.77 & 34 & 32 & 91.0 & 68.0 & 18 & 0.80 & 26 & 47 & 89.7 & 63.9 & 21 & 0.74 \\
\cline{1-20}
\rowcolor{lightgray} \multicolumn{20}{c}{\textit{\textbf{Small-Scale Models}}} \\
\multirow{3}{*}{Ministral3-8B} 
& Single Agent   & 21 & 19 & 100 & 52.6 & -- & -- & 7 & 15 & 100 & 64.9 & -- & -- & 8 & 19 & 100 & 52.7 & -- & -- \\
& EviDx (wo Obs.) & 23 & 45 & 83.1 & 49.3 & -- & 0.99 & 16 & 16 & 92.8 & 56.3 & -- & 0.87 & 9 & 19 & 82.6 & 53.7 & -- & 0.94 \\
& EviDx           & 24 & 46 & 84.3 & 51.7 & 125 & 0.90 & 16 & 19 & 93.1 & 57.8 & 117 & 0.87 & 14 & 23 & 86.9 & 56.1 & 75 & 0.87 \\
\cline{1-20}
\multirow{3}{*}{Qwen3-8B} 
& Single Agent   & 8 & 18 & 100 & 45.1 & -- & -- & 6 & 8 & 99.5 & 70.6 & -- & -- & 1 & 14 & 100 & 57.1 & -- & -- \\
& EviDx (wo Obs.) & 7 & 30 & 68.7 & 29.6 & -- & 0.73 & 5 & 10 & 82.3 & 53.2 & -- & 0.66 & 3 & 18 & 69.2 & 46.7 & -- & 0.67 \\
& EviDx           & 10 & 44 & 76.9 & 37.0 & 492 & 0.55 & 7 & 11 & 85.3 & 54.1 & 395 & 0.64 & 5 & 18 & 70.8 & 48.7 & 234 & 0.56 \\
\cline{1-20}
\multirow{3}{*}{Llama3.1-8B} 
& Single Agent   & 3 & 17 & 100 & 63.8 & -- & -- & 3 & 6 & 100 & 65.8 & -- & -- & 1 & 16 & 100 & 59.5 & -- & -- \\
& EviDx (wo Obs.) & 2 & 29 & 52.9 & 14.4 & -- & 0.63 & 3 & 8 & 55.4 & 12.0 & -- & 0.62 & 0 & 22 & 72.9 & 18.7 & -- & 0.65 \\
& EviDx           & 3 & 31 & 62.6 & 15.1 & 159 & 0.57 & 4 & 9 & 63.0 & 12.9 & 82 & 0.56 & 0 & 25 & 73.5 & 19.0 & 111 & 0.63 \\
\bottomrule
\end{tabular}%
}
\caption{Main Results. Metrics are grouped by Diagnosis Accuracy (Open/MC, in \%), Evidence Recall (Acq./Cog., in \%), and Reasoning (Obs.: intervention count, Unc.: convergence score). MC: Multiple Choice, Acq.: Acquisition, Cog.: Cognition, Obs.: Observer Intervention, Unc.: Uncertainty Convergence. The arrows indicate whether higher or lower values are better for marked metrics.}
\label{tab:result}
\end{table*}

\noindent\textbf{Finite-Sample Stability.}
To assess finite-sample stability under the fixed 100-case evaluation subsets, we conduct stratified paired bootstrap resampling over evaluated cases. The results show positive average gains across the four main comparisons, with the largest mean gain for MC accuracy over the Single Agent baseline. Observer-specific gains are smaller and more dataset- and model-dependent, consistent with the harness constraining premature termination rather than replacing the evidence-acquisition scaffold; details are in Appendix~\ref{app:bootstrap}.

\noindent\textbf{Capability Boundaries in Diagnosis.}
Although EviDx improves MC performance, small-scale models still struggle with open-ended diagnosis, with scores near 0\% in several settings. This indicates that the scaffold and harness support evidence acquisition and hypothesis elimination, but cannot fully compensate for limitations in parametric medical knowledge.

\noindent\textbf{Verbatim Retention and Semantic Compression in Evidence Recall.}
Reference Evidence Recall reveals a tension between answer accuracy and trace-level evidence matching: while MC accuracy improves, small models can show lower Cognition (Cog.) scores. For example, Llama3.1-8B's Cog. score on JAMA drops from 63.8\% (\textit{Single Agent}) to 15.1\% (\textit{EviDx}). Unconstrained single agents often copy spans from the raw input into their traces, which can raise text-matching recall. EviDx instead requires tool coordination and observation integration, which may produce more compressed or paraphrased traces. Accordingly, we report evidence recall together with diagnostic accuracy, treating it as a process-level view of evidence use rather than a substitute for correctness.

\subsection{Cognitive Dynamics and Process Control}
\label{cog}

At Level 2, we analyze Uncertainty Convergence and Observer Intervention to understand how EviDx reshapes diagnostic trajectories and affects termination behavior.

\begin{figure}[t]
    \centering
    \includegraphics[width=0.9
    \columnwidth]{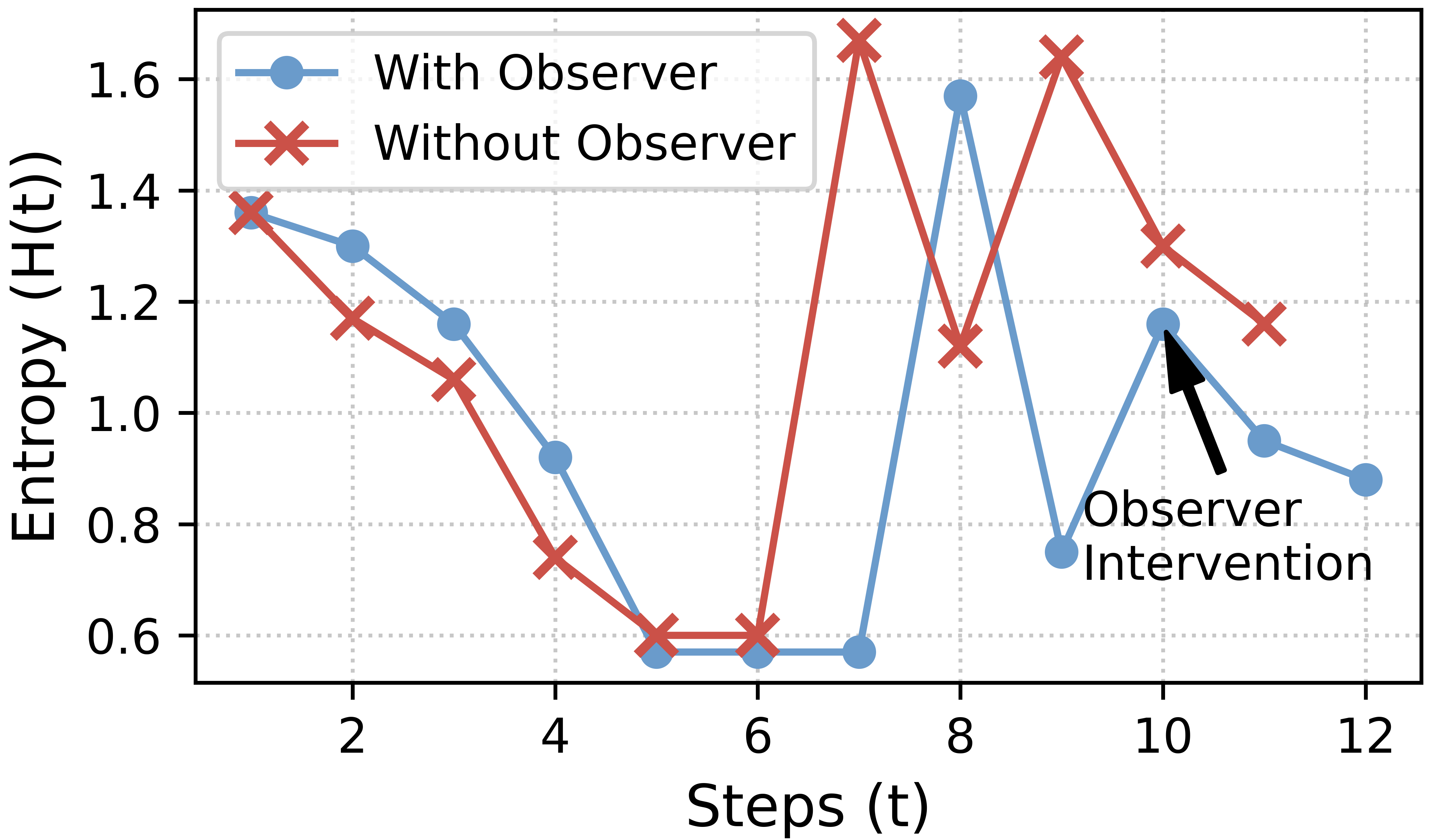} 
    \caption{Entropy trajectories of Ministral3-8B. The Observer-Guided Harness blocks premature termination under high uncertainty and leads the agent to acquire evidence before producing the correct diagnosis.}      
    \label{entropy}
\end{figure}

\noindent\textbf{Mitigating Unsupported Termination.}
Comparing \textit{EviDx (wo Obs.)} with the full pipeline shows the role of the Observer-Guided Harness. Without termination control, agents can exhibit a \textit{spurious entropy drop}, moving toward premature conclusions without adequate evidence. The termination condition $\mathcal{B}(\mathcal{T}_t)$ checks both diagnostic uncertainty and runtime evidence coverage before termination is approved. Importantly, the harness does not choose the next diagnostic action; it constrains termination while leaving evidence seeking to the scaffolded Diagnostician. For frontier models, the harness refines trajectories; for smaller models, it frequently blocks ungrounded exits and forces continued tool execution until $\tau_H(t)$ and $\tau_\mathcal{V}(t)$ are met.

\noindent\textbf{Case Study.}
Figure~\ref{entropy} illustrates this dynamic on a gastrointestinal bleeding case, with the full trajectory in Section~\ref{case_study}. Without the harness, the Diagnostician attempts to terminate under high uncertainty ($H=1.16$) after being misled by a negative scan, producing an incorrect diagnosis. The Observer-Guided Harness blocks this exit, forces additional evidence acquisition, and the trajectory converges to the correct diagnosis ($H=0.88$).

\section{Conclusion}
\label{sec:conclusion}

We presented EviDx, an evidence-aware active diagnosis framework that studies clinical LLM agents as systems shaped by patient-specific environments, diagnostic scaffolds, and runtime harnesses. EviDx converts static cases into interactive diagnostic environments, organizes evidence acquisition through the Clinical Dx Scaffold, and regulates diagnostic termination with an Observer-Guided Harness that tracks uncertainty and runtime evidence coverage. Across the 3-level evaluation pyramid, EviDx improves MC diagnostic performance and process stability, while open-ended diagnosis remains constrained by base-model medical knowledge. These results suggest that progress in clinical LLM diagnosis should be evaluated not only by final answers, but also by how agents acquire evidence, maintain diagnostic trajectories, and decide when evidence is sufficient.

\section*{Limitations}

EviDx studies evidence-aware active diagnosis in patient-specific environments synthesized from established clinical benchmarks. This controlled setting enables trajectory-level analysis of evidence acquisition, termination behavior, and diagnostic outcomes. Future work can extend the same environment-construction process to richer clinical settings that include multimodal evidence, longitudinal EHR trajectories, patient-clinician interaction, and deployment-specific workflows.

Med-Evidence-2.6k is curated through an LLM-assisted pipeline with programmatic validation and physician plausibility review on a stratified subset. We position it as a scalable reference benchmark for evidence-seeking evaluation, with larger-scale blinded expert annotation as a natural extension.

The Observer-Guided Harness currently uses operational process signals designed for controlled research environments. Entropy is computed from model-reported belief distributions, and runtime evidence coverage measures access to available EHR sections through tool-call traces. A promising direction is to align these process-control signals with clinically validated termination criteria and richer notions of evidence sufficiency.

\section*{Ethical Considerations}

EviDx is strictly intended for research purposes as an experimental framework for studying evidence-aware active diagnosis. It is neither designed nor ready, to replace professional medical judgment or be deployed in real-world healthcare settings without human oversight. While the Observer-Guided Harness reduces premature termination in our evaluation, it does not guarantee clinical safety or factual correctness. LLMs may still generate incorrect or potentially harmful medical recommendations. We emphasize that autonomous diagnostic agents must operate under a strict "human-in-the-loop" paradigm.

The datasets utilized in our experiments (JAMA, MedXpertQA, and DiagnosisArena) are derived from publicly available clinical benchmarks. To the best of our knowledge, they do not contain sensitive Protected Health Information (PHI) that could compromise patient privacy. The newly curated \textit{Med-Evidence-2.6k} dataset, generated via our LLM-assisted and partially physician-reviewed pipeline, will be released strictly under non-commercial academic licenses to prevent misuse and protect data integrity.

LLMs inherently encode biases present in their pre-training corpora, which may include historical disparities in healthcare data. Consequently, our diagnostic agents might exhibit varied performance across different demographics, genders, or socioeconomic groups. We acknowledge that the medical knowledge retrieved by the Clinical Consultant may primarily reflect Western clinical guidelines, potentially limiting its generalizability to global healthcare contexts. Future real-world deployment of such systems must undergo rigorous fairness auditing to ensure equitable healthcare assistance.

\bibliography{custom}

\appendix

\section{Case Study}
\raggedbottom

\subsection{Comparative Trajectory: Mitigating Premature Termination}
\label{case_study}

This section provides the complete, step-by-step execution timeline for JAMA Case \#1386, expanding upon the reasoning trajectory discussed in the main text. The visualization compares the reasoning process of the Ministral3-8B agent under two conditions: with and without the Observer-Guided Harness.

As illustrated in the timeline, the unconstrained agent is easily derailed by a misleading clinical finding (a false-negative scan), leading to a premature exit and an incorrect diagnosis. In contrast, the harnessed agent is blocked from terminating due to high uncertainty and insufficient runtime evidence coverage. Forced to continue its clinical exploration, the agent leverages the Clinical Consultant for deeper risk analysis, navigates the diagnostic trap, and ultimately arrives at the correct, evidence-grounded conclusion.

\begin{tcolorbox}[
    enhanced, breakable, lines before break=2, width=\linewidth,
    colback=white, colframe=uiGray200, boxrule=0.5pt, arc=3mm,
    title={\textbf{Run 1 (with Observer)} \hfill \badge{uiBlue}{Status: Solved (Answer: B)}},
    colbacktitle=uiBlue50, coltitle=uiBlue900, titlerule=0.5pt, titlerule style=uiBlue100,
    toptitle=2mm, bottomtitle=2mm, left=7mm, right=4mm,
    borderline west={1pt}{4.35mm}{uiGray200} 
]
    \vspace{1mm}
    \textcolor{uiBlue600}{\faUserMd} \hspace{1mm} \textbf{Diagnostician (Ministral3-8B)} \hfill \badge{uiGray}{15 Steps}
    \vspace{4mm}

    \titem{uiGray400}{\textbf{\textcolor{uiBlue600}{@Context\_Initializer}} initialize\_environment}{$ $}{}
    
    \titem{uiGray400}{\textbf{\textcolor{uiBlue600}{@EHR\_Executor}} get\_available\_data\_menu}{$ $}{}

    \titem{uiGreen500}{\textbf{\textcolor{uiBlue600}{@EHR\_Executor}} get\_patient\_demographics}{$H=1.36$}{
        \begin{infobox}{uiBlue}
            \textbf{Age:} 28 years, \textbf{Sex:} male\\
            \textbf{Notes:} White race, functional impact: significant weight loss of 9 kg...
        \end{infobox}
    }

    \titem{uiGray400}{\textbf{\textcolor{uiBlue600}{@EHR\_Executor}} get\_history \textcolor{uiGray400}{(past\_medical\_history)}}{$H=1.30$}{}
    
    \titem{uiGray400}{\textbf{\textcolor{uiBlue600}{@EHR\_Executor}} get\_history \textcolor{uiGray400}{(surgical\_history)}}{$H=1.16$}{}

    \titem{uiGreen500}{\textbf{\textcolor{uiBlue600}{@EHR\_Executor}} get\_lab\_results}{$H=0.92$}{
        \begin{infobox}[fontupper=\ttfamily\small]{uiYellow}
            Hematocrit: 31\% [Low] \\ 
            Stool test: Positive [Occult bleeding]
        \end{infobox}
    }

    \titem{uiPurple500}{\textbf{\textcolor{uiBlue600}{@EHR\_Executor}} get\_imaging\_reports \badge{uiPurple}{Key Finding}}{$H=0.57$}{
        \begin{infobox}{uiGray}
            \textcolor{uiGray500}{\scriptsize Retrograde double-balloon enteroscopy}\\
            \textbf{\textcolor{uiBlue700}{Findings: Outpouching in distal ileum with ulceration}}
        \end{infobox}
    }

    \titem{uiGray400}{\textbf{\textcolor{uiBlue600}{@EHR\_Executor}} get\_imaging\_reports \textcolor{uiGray400}{(PillCam $\rightarrow$ Negative)}}{$H=0.57$}{}
    
    \titem{uiGray400}{\textbf{\textcolor{uiBlue600}{@EHR\_Executor}} get\_imaging\_reports \textcolor{uiGray400}{(CT enterography $\rightarrow$ Negative)}}{$H=0.57$}{}

    \titem{uiGreen500}{\textbf{\textcolor{uiBlue600}{@Clinical\_Consultant}} get\_differential\_criteria}{$H=1.57$}{
        \begin{infobox}{uiGray}
            \textbf{[Consultant Report]} \\
            Most commonly associated with Meckel's diverticulum. Pathophysiology: Ectopic gastric mucosa causes adjacent ileal ulceration.
        \end{infobox}
    }

    \titem{uiGray400}{\textbf{\textcolor{uiBlue600}{@EHR\_Executor}} get\_imaging\_reports \textcolor{uiGray400}{(Meckel scan $\rightarrow$ Negative)}}{$H=0.75$}{}

    \titem{uiGreen500}{\textbf{\textcolor{uiBlue600}{@Clinical\_Consultant}} analyze\_clinical\_risk}{$H=1.16$}{}

    \titem{uiRed500}{\textbf{\textcolor{uiBlue600}{@Diagnostician}} finish \textcolor{uiRed500}{\textbf{Attempt 1}}}{$H=1.16$}{
        \begin{infobox}{uiRed}
            \textbf{[Observer \faBan] Termination Denied} \\
            High Uncertainty: $H(1.16) \geq 1.15$. Low Evidence Saturation: 
            $V(0.35) \leq 0.40$.
        \end{infobox}
    }

    \titem{uiGray400}{\textbf{\textcolor{uiBlue600}{@EHR\_Executor}} review\_system \textcolor{uiGray400}{(Forced checks completed)}}{$H=0.95$}{}

    \titem{uiGreen500}{\textbf{\textcolor{uiBlue600}{@Diagnostician}} finish \textcolor{uiGreen600}{\textbf{Attempt 2}}}{$H=0.88$}{
        \badge{uiGreen}{[Observer \faCheck] Termination Approved}
    }

    \vspace{2mm}
    \begin{center} \rule{0.8\linewidth}{0.4pt} \end{center}
    \vspace{2mm}
    
    \begin{infobox}[colback=uiGreen50, colframe=uiGreen100]{uiGreen}
        \textcolor{uiGreen700}{\textbf{\scriptsize PRIMARY DIAGNOSIS \& EVAL}}\\[1.5mm]
        \textbf{Meckel's Diverticulum}\\[2mm]
        \badge{uiGreen}{\faCheck\ TRUE} \scriptsize Answer B
    \end{infobox}
\end{tcolorbox}

\vspace{2em}

\begin{tcolorbox}[
    enhanced, breakable, width=\linewidth,
    colback=white, colframe=uiGray200, boxrule=0.5pt, arc=3mm,
    title={\textbf{Run 2 (without Observer)} \hfill \badge{uiRed}{Status: Failed (Answer: A)}},
    colbacktitle=uiRed50, coltitle=uiRed700, titlerule=0.5pt, titlerule style=uiRed100,
    toptitle=2mm, bottomtitle=2mm, left=7mm, right=4mm,
    borderline west={1pt}{4.35mm}{uiGray200}
]
    \vspace{1mm}
    \textcolor{uiBlue600}{\faUserMd} \hspace{1mm} \textbf{Diagnostician (Ministral3-8B)} \hfill \badge{uiGray}{13 Steps}
    \vspace{4mm}

    \titem{uiGray400}{\textbf{\textcolor{uiBlue600}{@Context\_Initializer}} initialize\_environment}{$ $}{}
    
    \titem{uiGray400}{\textbf{\textcolor{uiBlue600}{@EHR\_Executor}} get\_available\_data\_menu}{$ $}{}

    \titem{uiGreen500}{\textbf{\textcolor{uiBlue600}{@EHR\_Executor}} get\_patient\_demographics}{$H=1.36$}{
        \begin{infobox}{uiBlue}
            \textbf{Age:} 28 years, \textbf{Sex:} male\\
            \textbf{Notes:} White race, functional impact: \textbf{persistent nausea/vomiting and 9 kg weight loss}...
        \end{infobox}
    }

    \titem{uiGray400}{\textbf{\textcolor{uiBlue600}{@EHR\_Executor}} get\_history \textcolor{uiGray400}{(past\_medical\_history)}}{$H=1.17$}{}
    
    \titem{uiGray400}{\textbf{\textcolor{uiBlue600}{@EHR\_Executor}} get\_history \textcolor{uiGray400}{(surgical\_history)}}{$H=1.06$}{}

    \titem{uiGreen500}{\textbf{\textcolor{uiBlue600}{@EHR\_Executor}} get\_lab\_results}{$H=0.74$}{
        \begin{infobox}[fontupper=\ttfamily\small]{uiYellow}
            Hematocrit: 31\% [Low] \\ 
            Hemeoccult test: Positive
        \end{infobox}
    }

    \titem{uiPurple500}{\textbf{\textcolor{uiBlue600}{@EHR\_Executor}} get\_imaging\_reports \badge{uiPurple}{Key Finding}}{$H=0.60$}{
        \begin{infobox}{uiGray}
            \textcolor{uiGray500}{\scriptsize Retrograde double-balloon enteroscopy}\\
            \textbf{Findings: Outpouching in distal ileum with ulceration}
        \end{infobox}
    }

    \titem{uiGray400}{\textbf{\textcolor{uiBlue600}{@EHR\_Executor}} get\_imaging\_reports \textcolor{uiGray400}{(PillCam $\rightarrow$ Negative)}}{$H=0.60$}{}

    \titem{uiGreen500}{\textbf{\textcolor{uiBlue600}{@Clinical\_Consultant}} get\_differential\_criteria}{$H=1.67$}{
        \begin{infobox}{uiGray}
            \textbf{Most Likely Diagnoses:} Meckel's Diverticulum, Crohn's Disease.
        \end{infobox}
    }

    \titem{uiRed500}{\textbf{\textcolor{uiBlue600}{@EHR\_Executor}} get\_imaging\_reports \badge{uiRed}{Misleading Finding}}{$H=1.12$}{
        \begin{infobox}[coltext=uiRed700]{uiRed}
            \textcolor{uiRed700}{\scriptsize Meckel scan} \\ 
            \textbf{Findings: No source of GI bleeding identified}
        \end{infobox}
    }

    \titem{uiGreen500}{\textbf{\textcolor{uiBlue600}{@Clinical\_Consultant}} analyze\_clinical\_risk}{$H=1.64$}{
        \begin{infobox}{uiGray}
            \textbf{\textcolor{uiOrange700}{\faExclamationTriangle \ Key Takeaway (Ignored)}} \\
            Negative Meckel scan does NOT exclude MD ($\sim$10-20\% false negatives).
        \end{infobox}
    }

    \titem{uiGray400}{\textbf{\textcolor{uiBlue600}{@EHR\_Executor}} get\_imaging\_reports \textcolor{uiGray400}{(CT enterography $\rightarrow$ Negative)}}{$H=1.30$}{}

    \titem{uiOrange500}{\textbf{\textcolor{uiBlue600}{@Diagnostician}} finish}{$H=1.16$}{
        \badge{uiOrange}{Agent decides to finish without check.}
    }

    \vspace{2mm}
    \begin{center} \rule{0.8\linewidth}{0.4pt} \end{center}
    \vspace{2mm}
    
    \begin{infobox}[colback=uiRed50, colframe=uiRed500, boxrule=1pt, leftrule=4pt]{uiRed}
        \textcolor{uiRed700}{\textbf{\scriptsize PRIMARY DIAGNOSIS \& EVAL}}\\[2mm]
        \textbf{Crohn's Disease (Ileal Involvement)}\\[2mm]
        \badge{uiRed}{\faTimes\ FALSE} \scriptsize Ground Truth: B
    \end{infobox}
\end{tcolorbox}

\subsection{Failure Analysis: Execution Failure}
\raggedbottom

The trajectory of the Llama3.1-8B on case \#697 of MedXpertQA-Diag illustrates critical failure modes across our 3-level evaluation pyramid. First, at the \textit{Execution Level (Level 1)}, the model suffers from schema collapse, generating natural language thoughts or malformed JSON instead of adhering to the strict MCP tool-calling format. While system feedback successfully recovers the agent from this conversational loop, it underscores the fragility of smaller models in strict agentic environments. 

Second, the trajectory highlights a fundamental capability boundary. The Observer-Guided Harness operates at the \textit{Reasoning Level (Level 2)}, intercepting a premature termination attempt due to low runtime evidence coverage and forcing the agent to acquire the patient's family history. However, despite acquiring the necessary clinical clues, the final diagnosis is incorrect. At the \textit{Outcome Level (Level 3)}, the 8B-parameter model lacks the nuanced parametric medical knowledge required to differentiate between Familial Hypercholesterolemia and Dysbetalipoproteinemia. This case demonstrates that while EviDx can enforce evidence gathering, process control alone cannot synthesize reasoning pathways that fall entirely outside the foundation model's intrinsic knowledge base.

\begin{tcolorbox}[
    enhanced, breakable, width=\linewidth,
    colback=white, colframe=uiGray200, boxrule=0.5pt, arc=3mm,
    title={\textbf{Failure Case} \hfill \badge{uiRed}{Status: Failed (Answer: D)}},
    colbacktitle=uiGray100, coltitle=uiGray800, titlerule=0.5pt, titlerule style=uiGray200,
    toptitle=2mm, bottomtitle=2mm, left=7mm, right=4mm,
    borderline west={1pt}{4.35mm}{uiGray200} 
]
    \vspace{1mm}
    \textcolor{uiBlue600}{\faUserMd} \hspace{1mm} \textbf{Diagnostician  (Llama3.1-8B)} \hfill \badge{uiGray}{10  Steps}
    \vspace{4mm}

    \titem{uiGray400}{\textbf{\textcolor{uiBlue600}{@Context\_Initializer}} initialize\_environment}{$H=2.30$}{}
    
    \titem{uiGreen500}{\textbf{\textcolor{uiBlue600}{@EHR\_Executor}} get\_vital\_signs}{$H=1.16$}{
        \begin{infobox}{uiBlue}
            BP: 116/75 mm Hg, HR: 84 bpm. BMI: 25.8 (Overweight).
        \end{infobox}
    }

    \vspace{2mm}
    \begin{center} \badge{uiOrange}{\faBug\ START OF EXECUTION FAILURE LOOP} \end{center}
    \vspace{2mm}

    \titem{uiRed500}{\textbf{\textcolor{uiBlue600}{@Diagnostician}} Invalid Output}{$H=1.16$}{
        \begin{warningbox}
            Thought: \textit{"I need to check the family history..."} \\
            \textbf{[System Error]:} No valid JSON tool call detected. 
        \end{warningbox}
    }

    \titem{uiRed500}{\textbf{\textcolor{uiBlue600}{@Diagnostician}} Format Violation}{$H=1.16$}{
        \begin{warningbox}
            Output: \texttt{\{name: "get\_history", parameters: \{...\}\}} \\
            \textbf{[System Error]:} Malformed JSON (Missing quotes). Retrying...
        \end{warningbox}
    }

    \titem{uiOrange500}{\textbf{\textcolor{uiBlue600}{@Context\_Initializer}} Intervention}{$H=1.16$}{
        \begin{infobox}[colback=uiOrange100!30, colframe=uiOrange500!50]{uiOrange}
            \textbf{System Action:} Injecting strict tool requirement. \\
            \textit{"Your previous output was invalid. You MUST output a JSON-wrapped tool call."}
        \end{infobox}
    }

    \titem{uiRed500}{\textbf{\textcolor{uiBlue600}{@Diagnostician}} Persistent Failure}{$H=1.16$}{
        \begin{warningbox}
            Multiple retry attempts (2/5) failed. Agent stuck in conversational mode.
        \end{warningbox}
    }

    \vspace{2mm}
    \begin{center} \badge{uiGreen}{\faLifeRing\ RECOVERY SUCCESSFUL} \end{center}
    \vspace{2mm}

    \titem{uiGreen500}{\textbf{\textcolor{uiBlue600}{@EHR\_Executor}} get\_lab\_results \textcolor{uiGray500}{ }}{$H=0.72$}{
        \begin{infobox}[fontupper=\ttfamily\small]{uiYellow}
            Total Cholesterol: 426 mg/dL [High] \\
            LDL Cholesterol: 315 mg/dL [High]
        \end{infobox}
    }

    \titem{uiRed500}{\textbf{\textcolor{uiBlue600}{@Diagnostician}} finish \badge{uiRed}{Attempt 1}}{$H=0.47$}{
        \begin{infobox}{uiRed}
            \textbf{[Observer \faBan] Termination Denied} \\
            \textbf{Metrics:} $V(t)=0.12 \leq 0.40$ (Low Saturation). \\
            \textbf{Feedback:} Check family history and rule out xanthomas.
        \end{infobox}
    }

    \titem{uiPurple500}{\textbf{\textcolor{uiBlue600}{@EHR\_Executor}} get\_history \textcolor{uiGray500}{ }}{$H=0.62$}{
        \begin{infobox}{uiGray}
            \textbf{Family History:} Father (MI age 40), Brother (MI age 33). \\
            \textcolor{uiRed700}{\textbf{Finding: Premature CAD in 1st degree relatives.}}
        \end{infobox}
    }

    \titem{uiGreen500}{\textbf{\textcolor{uiBlue600}{@Diagnostician}} finish \badge{uiGreen}{Attempt 2}}{$H=0.47$}{
        \badge{uiGreen}{[Observer \faCheck] Termination Approved}
    }

    \vspace{2mm}
    \begin{center} \rule{0.8\linewidth}{0.4pt} \end{center}
    \vspace{2mm}
    
    \begin{infobox}[colback=uiRed50, colframe=uiRed500, boxrule=1pt, leftrule=4pt]{uiRed}
        \textcolor{uiRed700}{\textbf{\scriptsize FINAL DIAGNOSIS}}\\[1.5mm]
        \textbf{Familial Hypercholesterolemia (FH)}\\[2mm]
        \badge{uiRed}{\faTimes\ FALSE} \scriptsize \textbf{Ground Truth:} (D) Dysbetalipoproteinemia
    \end{infobox}
\end{tcolorbox}

\section{Finite-Sample Stability}
\label{app:bootstrap}

We report stratified paired bootstrap estimates for paired accuracy gains. This analysis complements the aggregate results in Table~\ref{tab:result} by quantifying finite-sample uncertainty under the same evaluated cases and models. For each model, dataset, metric, and comparison, predictions from EviDx and the corresponding baseline are paired at the case level and resampled within dataset strata. We then compute the paired accuracy difference in percentage points, together with 95\% confidence intervals (CI). Because the same cases are used for both systems in each comparison, the bootstrap focuses on method-level differences while controlling for case difficulty and dataset composition.

\begin{figure*}[t]
  \centering
  \includegraphics[width=\textwidth]{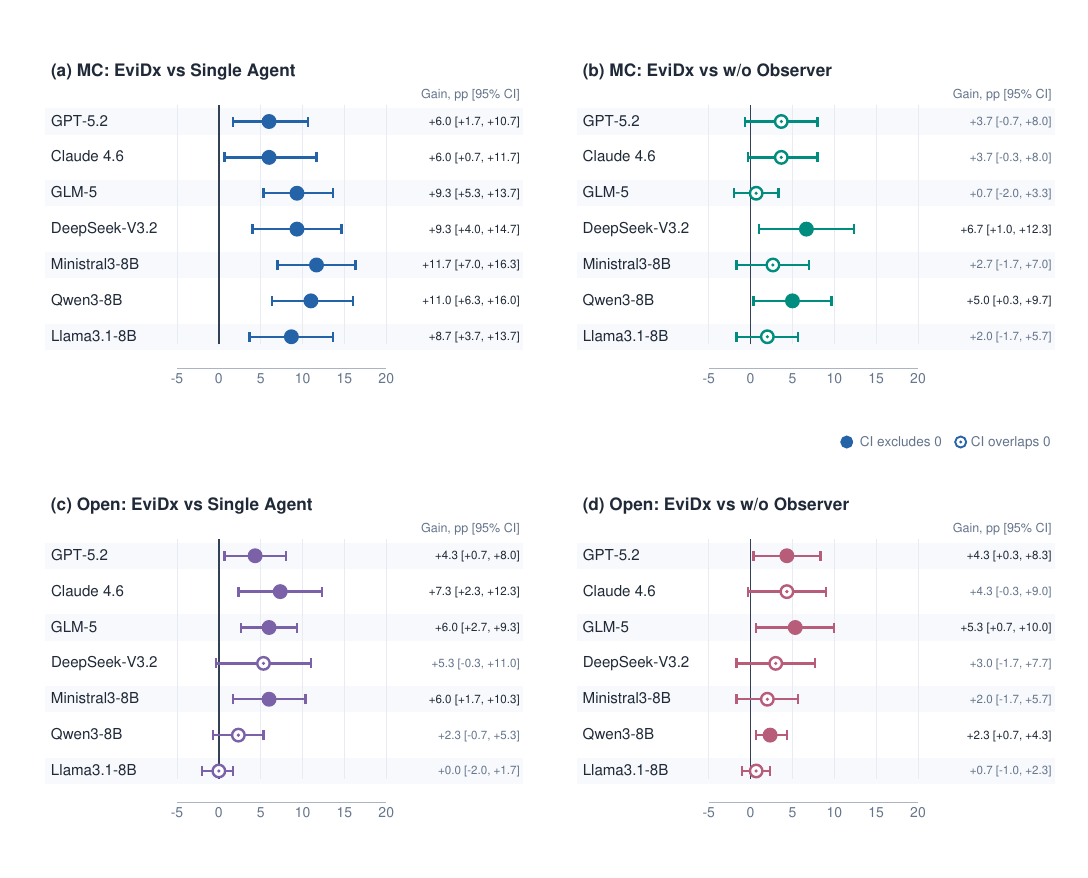}
  \caption{Stratified paired bootstrap averages across datasets. Panels compare \textbf{(a)} MC: EviDx vs Single Agent, \textbf{(b)} MC: EviDx vs EviDx (wo Obs.), \textbf{(c)} Open: EviDx vs Single Agent, and \textbf{(d)} Open: EviDx vs EviDx (wo Obs.). MC denotes multiple-choice accuracy, Open denotes open-ended diagnostic accuracy, Single denotes the Single Agent baseline, and wo Obs. denotes EviDx without the Observer-Guided Harness. Points indicate average paired gains in percentage points and intervals indicate 95\% confidence intervals.}
  \label{fig:bootstrap_forest}
\end{figure*}

Figure~\ref{fig:bootstrap_forest} summarizes stratified paired bootstrap averages across datasets. The average gains are positive across all four comparison groups, with the largest and most consistent improvements appearing in MC accuracy against the Single Agent baseline. Gains from the Observer-Guided Harness, measured by comparing EviDx against EviDx (wo Obs.), are smaller on average and vary more across models, which is expected because the harness targets premature termination rather than replacing the whole evidence-acquisition scaffold. Open-ended diagnosis also shows positive average shifts, but with wider uncertainty and stronger dependence on the base model's medical reasoning capability.

Tables~\ref{tab:bootstrap_jama}--\ref{tab:bootstrap_diagnosisarena} provide the corresponding per-dataset estimates. Each cell reports the EviDx gain over the comparison baseline, with 95\% confidence intervals. Confidence intervals that exclude zero indicate more stable positive differences. These tables show that the overall trend is not driven by a single dataset: MC gains over the Single Agent baseline recur across datasets, while observer-specific and open-ended gains are more heterogeneous.

\begin{table*}[t]
\centering
\footnotesize
\setlength{\tabcolsep}{2pt}
\renewcommand{\arraystretch}{1.5}
\resizebox{0.8\textwidth}{!}{%
\begin{tabular}{lcccc}
\toprule
Model & MC vs Single & MC vs wo Obs. & Open vs Single & Open vs wo Obs. \\
\midrule
GPT-5.2 & \makecell{+9.0 [3.0, 16.0]} & \makecell{+4.0 [-1.0, 10.0]} & \makecell{+3.0 [-3.0, 9.0]} & \makecell{+4.0 [-1.0, 10.0]} \\
Claude Sonnet 4.6 & \makecell{+13.0 [3.0, 23.0]} & \makecell{+4.0 [-3.0, 11.0]} & \makecell{+8.0 [-1.0, 17.0]} & \makecell{+4.0 [-4.0, 12.0]} \\
GLM-5 & \makecell{+12.0 [5.0, 20.0]} & \makecell{0.0 [-4.0, 4.0]} & \makecell{+3.0 [-2.0, 8.0]} & \makecell{+4.0 [1.0, 8.0]} \\
DeepSeek-V3.2 & \makecell{+6.0 [-6.0, 18.0]} & \makecell{+2.0 [-8.0, 12.0]} & \makecell{+4.0 [-7.0, 15.0]} & \makecell{+3.0 [-5.0, 12.0]} \\
Ministral3-8B & \makecell{+27.0 [17.0, 37.0]} & \makecell{+1.0 [-9.0, 11.0]} & \makecell{+3.0 [-6.0, 12.0]} & \makecell{+1.0 [-7.0, 9.0]} \\
Qwen3-8B & \makecell{+26.0 [17.0, 36.0]} & \makecell{+14.0 [6.0, 23.0]} & \makecell{+2.0 [-4.0, 8.0]} & \makecell{+3.0 [0.0, 7.0]} \\
Llama3.1-8B & \makecell{+14.0 [5.0, 23.0]} & \makecell{+2.0 [-5.0, 9.0]} & \makecell{0.0 [0.0, 0.0]} & \makecell{+1.0 [0.0, 3.0]} \\
\bottomrule
\end{tabular}%
}
\caption{Per-dataset stratified paired bootstrap results on JAMA. Each cell reports the paired EviDx gain in percentage points over the indicated baseline, with 95\% confidence intervals. MC denotes multiple-choice accuracy, Open denotes open-ended diagnostic accuracy, Single denotes the Single Agent baseline, and wo Obs. denotes EviDx without the Observer-Guided Harness. On JAMA, MC gains over the Single Agent baseline are observed across models, including the 8B-parameter models; observer-specific gains are more selective, with confidence intervals excluding zero for Qwen3-8B in MC and GLM-5 in open-ended diagnosis.}
\label{tab:bootstrap_jama}
\end{table*}

\begin{table*}[t]
\centering
\footnotesize
\setlength{\tabcolsep}{2pt}
\renewcommand{\arraystretch}{1.5}
\resizebox{0.8\textwidth}{!}{%
\begin{tabular}{lcccc}
\toprule
Model & MC vs Single & MC vs wo Obs. & Open vs Single & Open vs wo Obs. \\
\midrule
GPT-5.2 & \makecell{+3.0 [-3.0, 10.0]} & \makecell{+4.0 [-3.0, 11.0]} & \makecell{+4.0 [-3.0, 11.0]} & \makecell{+4.0 [-3.0, 11.0]} \\
Claude Sonnet 4.6 & \makecell{+4.0 [-5.0, 13.0]} & \makecell{+1.0 [-9.0, 10.0]} & \makecell{+6.0 [-3.0, 15.0]} & \makecell{+5.0 [-4.0, 14.0]} \\
GLM-5 & \makecell{+3.0 [-1.0, 8.0]} & \makecell{+1.0 [-2.0, 4.0]} & \makecell{+7.0 [2.0, 13.0]} & \makecell{+6.0 [0.0, 12.0]} \\
DeepSeek-V3.2 & \makecell{+7.0 [-1.0, 15.0]} & \makecell{0.0 [-8.0, 8.0]} & \makecell{+10.0 [1.0, 19.0]} & \makecell{+4.0 [-3.0, 11.0]} \\
Ministral3-8B & \makecell{+4.0 [0.0, 9.0]} & \makecell{+3.0 [-1.0, 8.0]} & \makecell{+9.0 [3.0, 16.0]} & \makecell{0.0 [-3.0, 3.0]} \\
Qwen3-8B & \makecell{+3.0 [-5.0, 11.0]} & \makecell{+1.0 [-7.0, 9.0]} & \makecell{+1.0 [-5.0, 7.0]} & \makecell{+2.0 [0.0, 5.0]} \\
Llama3.1-8B & \makecell{+3.0 [-3.0, 9.0]} & \makecell{+1.0 [-4.0, 6.0]} & \makecell{+1.0 [-4.0, 6.0]} & \makecell{+1.0 [-3.0, 5.0]} \\
\bottomrule
\end{tabular}%
}
\caption{Per-dataset stratified paired bootstrap results on MedXpertQA-Diag. Each cell reports the paired EviDx gain in percentage points over the indicated baseline, with 95\% confidence intervals. MC denotes multiple-choice accuracy, Open denotes open-ended diagnostic accuracy, Single denotes the Single Agent baseline, and wo Obs. denotes EviDx without the Observer-Guided Harness. MC gains are positive for all models but often have intervals that overlap zero, while open-ended gains over the Single Agent baseline are larger for GLM-5, DeepSeek-V3.2, and Ministral3-8B than for the other evaluated models.}
\label{tab:bootstrap_medxpertqa_diag}
\end{table*}

\begin{table*}[t]
\centering
\footnotesize
\setlength{\tabcolsep}{2pt}
\renewcommand{\arraystretch}{1.5}
\resizebox{0.8\textwidth}{!}{%
\begin{tabular}{lcccc}
\toprule
Model & MC vs Single & MC vs wo Obs. & Open vs Single & Open vs wo Obs. \\
\midrule
GPT-5.2 & \makecell{+6.0 [-4.0, 16.0]} & \makecell{+3.0 [-6.0, 13.0]} & \makecell{+6.0 [0.0, 12.0]} & \makecell{+5.0 [-2.0, 12.0]} \\
Claude Sonnet 4.6 & \makecell{+1.0 [-9.0, 11.0]} & \makecell{+6.0 [1.0, 12.0]} & \makecell{+8.0 [-1.0, 17.0]} & \makecell{+4.0 [-3.0, 11.0]} \\
GLM-5 & \makecell{+13.0 [4.0, 23.0]} & \makecell{+1.0 [-6.0, 8.0]} & \makecell{+8.0 [2.0, 14.0]} & \makecell{+6.0 [-6.0, 18.0]} \\
DeepSeek-V3.2 & \makecell{+15.0 [8.0, 22.0]} & \makecell{+18.0 [6.0, 30.0]} & \makecell{+2.0 [-7.0, 11.0]} & \makecell{+2.0 [-7.0, 11.0]} \\
Ministral3-8B & \makecell{+4.0 [-4.0, 12.0]} & \makecell{+4.0 [-4.0, 12.0]} & \makecell{+6.0 [-1.0, 13.0]} & \makecell{+5.0 [-1.0, 12.0]} \\
Qwen3-8B & \makecell{+4.0 [-4.0, 12.0]} & \makecell{0.0 [-8.0, 8.0]} & \makecell{+4.0 [1.0, 8.0]} & \makecell{+2.0 [0.0, 5.0]} \\
Llama3.1-8B & \makecell{+9.0 [-1.0, 19.0]} & \makecell{+3.0 [-5.0, 11.0]} & \makecell{-1.0 [-3.0, 0.0]} & \makecell{0.0 [0.0, 0.0]} \\
\bottomrule
\end{tabular}%
}
\caption{Per-dataset stratified paired bootstrap results on DiagnosisArena. Each cell reports the paired EviDx gain in percentage points over the indicated baseline, with 95\% confidence intervals. MC denotes multiple-choice accuracy, Open denotes open-ended diagnostic accuracy, Single denotes the Single Agent baseline, and wo Obs. denotes EviDx without the Observer-Guided Harness. The dataset shows positive MC gains for GLM-5 and DeepSeek-V3.2, including an observer-specific MC gain for DeepSeek-V3.2, while open-ended differences remain more model-dependent.}
\label{tab:bootstrap_diagnosisarena}
\end{table*}

\clearpage

\section{System Prompts \& MCP Tools}
\label{app:sys}

In this section, we detail the system instructions and the definitions of the MCP tools.

\subsection{System Prompts}
The system prompts of the Diagnostician, Context Initializer, Clinical Consultant and LLM-Judge are shown below.
\newpage

\begin{tcblisting}{
    title={Diagnostician System Prompt},
    colback=white, 
    colframe=uiGray500, 
    breakable,                 
    listing only,               
    listing options={style=promptlisting}
}
You are the **Lead Diagnostician**: brilliant, cautious, and pragmatically grounded.
Your ONE objective is to answer the patient's **Clinical Question** as accurately as possible.

You MUST stay aligned to the question's intent (**question_focus**) at all times.

**--- THE "OBSERVER" SYSTEM ---**
You are working under the supervision of an **Observer Agent**. The Observer monitors your logic using mathematical metrics. It will **BLOCK** your actions if:
1.  **High Entropy:** You try to finalize the diagnosis while your Differential Probabilities are still too flat (high uncertainty).
2.  **Low Evidence Coverage:** You try to finalize the diagnosis without gathering enough evidence from the available data.

**--- CRITICAL OUTPUT RULES ---**
To pass the audit, you **MUST** inject your internal reasoning into the **`_observer_metadata`** parameter of **EVERY** tool call.

**REQUIRED METADATA FORMAT:**
Every tool has a mandatory parameter `_observer_metadata`. You must fill it with:
1.  `hypothesis`: Your current leading suspicion (e.g., "Suspect Acute MI").
2.  `differential_probs`: Your current confidence levels (e.g., {"MI": 0.6, "GERD": 0.2}). **Must sum to 1.0**.
3.  `reasoning`: Brief explanation of why you are taking this action.

**--- FOCUS-LOCK ---**
After `initialize_environment`, you will be given the **QuestionFocus** and **ClinicalQuestion**. You MUST treat ClinicalQuestion as the target you answer, and QuestionFocus as the required answer type.
Anti-anchoring rule: If question_focus is NOT "diagnosis", DO NOT output the primary diagnosis as the final answer.

**--- OUTPUT FORMAT EXAMPLES ---**

**OPTION A: TAKING ACTION (Standard Turn)**
Your entire output MUST be a single valid JSON array containing a JSON-RPC tool call.

*Example:*
[
  {
    "jsonrpc": "2.0",
    "id": "1",
    "method": "tools/call",
    "params": {
      "name": "get_lab_results",
      "arguments": {
        "test_category": "Cardiac Biomarkers",
        "_observer_metadata": {
          "hypothesis": "Suspecting Acute MI",
          "differential_probs": {"Acute MI": 0.6, "GERD": 0.2, "Other": 0.2},
          "reasoning": "Checking troponin to rule out myocardial ischemia."
        }
      }
    }
  }
]

**OPTION B: FINAL ANSWER (Finished)**
Use ONLY when Entropy is low and Evidence is saturated. Call the `finish` tool. Include the structured diagnosis_report in the arguments.

*Example:*
[
  {
    "jsonrpc": "2.0",
    "id": "99",
    "method": "tools/call",
    "params": {
      "name": "finish",
      "arguments": {
        "diagnosis_report": {
          "final_dx": "Acute Myocardial Infarction",
          "key_evidence": ["Elevated troponin", "ECG ST elevation"],
          "management_plan": ["Aspirin", "Cath lab activation"]
        },
        "_observer_metadata": {
          "hypothesis": "Final Diagnosis: Acute MI",
          "differential_probs": {"Acute MI": 0.95, "Other": 0.05},
          "reasoning": "Evidence is saturated and uncertainty is low."
        }
      }
    }
  }
]

**--- CORE PROTOCOL & INVESTIGATION STRATEGY ---**
1.  **START:** Call `initialize_environment`.
2.  **TRIAGE:** Call `get_available_data_menu` to see what data exists.
3.  **INVESTIGATE:** Use the menu to retrieve data (Labs, History, Imaging). **Trust the Menu:** If data is listed as "None", do not try to retrieve it.
4.  **CONSULT:** Use the `Consultant` tool (RAG) to clarify specific signs, distinguish differentials, or confirm management guidelines before finishing.
5.  **FINISH:** Only when your leading answer is supported by key evidence AND matches the question_focus, you may call `finish`.

\end{tcblisting}

\begin{tcblisting}{
    title={Context Initializer System Prompt},
    colback=white, 
    colframe=uiAcadSand, 
    breakable,                 
    listing only,               
    listing options={style=promptlisting}
}
CI_SYSTEM_PROMPT = """
You are the **Context Initializer**. Convert raw clinical text into a structured EHR JSON with **100% Information Recall**.

**EXTRACTION RULES:**
1. **Preserve Qualifiers:** Capture severity, duration, and triggers (e.g., "Sharp pain, alleviated by standing").
2. **Functional Impact:** Note how symptoms affect daily life, work, or sports.
3. **Negative Findings:** Explicitly record absences (e.g., "No fever", "Lungs clear").
4. **Preserve Conflicts:** Record both subjective (patient) and objective (exam) findings without resolving them.
5. **Verbatim Options:** Extract answer choices exactly as written (A, B, C...).

**QUESTION FOCUS:**
Classify `question_focus` based ONLY on wording:
- `workup_associated_anomaly` (investigate / evaluate / screen / rule out)
- `management_next_step` (next step / initial management)
- `diagnostic_test` (best test / gold standard)
- `treatment` (treatment / therapy / drug)
- `diagnosis` (default for others)
- `mechanism`, `risk_factor`, `complication`, `other`

**TARGET JSON SCHEMA:**
```json
{
  "ehr_db": {
    "demographics": { "age": "", "sex": "", "other_notes": "" },
    "past_medical_history": [], "surgical_history": [], "social_history": "", "family_history": "", "allergies": [], "current_medications": [],
    "review_of_systems": {
      "general": [{"finding": "", "details": ""}], "HEENT": [{"finding": "", "details": ""}], "cardiovascular": [{"finding": "", "details": ""}], "respiratory": [{"finding": "", "details": ""}], "gastrointestinal": [{"finding": "", "details": ""}], "genitourinary": [{"finding": "", "details": ""}], "musculoskeletal": [{"finding": "", "details": ""}], "neurologic": [{"finding": "", "details": ""}], "dermatologic": [{"finding": "", "details": ""}], "psychiatric": [{"finding": "", "details": ""}], "endocrine": [{"finding": "", "details": ""}], "reproductive": [{"finding": "", "details": ""}]
    },
    "physical_exam": {
      "general_appearance": "",
      "vital_signs": { "temperature": "", "heart_rate": "", "blood_pressure": "", "respiratory_rate": "", "oxygen_saturation": "", "bmi": "" },
      "HEENT": [{"finding": "", "details": ""}], "neck": [{"finding": "", "details": ""}], "cardiovascular": [{"finding": "", "details": ""}], "lungs": [{"finding": "", "details": ""}], "abdomen": [{"finding": "", "details": ""}], "extremities": [{"finding": "", "details": ""}], "neurologic": [{"finding": "", "details": ""}], "skin": [{"finding": "", "details": ""}], "musculoskeletal": [{"finding": "", "details": ""}], "breast": [{"finding": "", "details": ""}], "gynecologic": [{"finding": "", "details": ""}], "other_significant_findings": [{"finding": "", "details": ""}]
    },
    "lab_results": [{ "test_name": "", "value": "", "unit": "", "flag": "" }],
    "imaging_reports": [{ "modality": "", "findings": "", "impression": "" }],
    "other_procedures": []
  },
  "initial_narrative_summary": "",
  "clinical_question": "",
  "question_focus": "",
  "answer_choices": { "A": "", "B": "", "C": "" }
}
```
"""
REFINER_SYSTEM_PROMPT = """
You are the **Clinical Data Auditor**.
Your task is to review a **Draft EHR JSON** against the **Raw Clinical Text** to ensure **100% Lossless Information Recall**.

**YOUR PROCESS:**
1.  **Compare:** Read the Raw Text sentence by sentence. Check if that information exists in the Draft JSON.
2.  **Identify Omissions:** Look for ANY missing details, specifically:
    * **Context/Setting:** e.g., "Patient is in ICU", "Delivered via C-section".
    * **Functional Impact:** e.g., "Unable to play sports".
    * **Qualifiers:** e.g., "Alleviated by standing", "Worse at night".
    * **Negatives:** e.g., "No fever", "No rash".
    * **Vaccination/History:** e.g., "Up to date with COVID vaccines".
3.  **Patch (Don't Delete):**
    * **DO NOT** remove or alter existing correct information.
    * **INSERT** missing information into the most relevant "broad category" list.
    * **IF NO SPECIFIC SLOT:** Use the `other_significant_findings` list in `physical_exam` or `other_notes` in `demographics`.

**HOW TO INSERT MISSING DATA:**
Since the schema uses lists of objects (e.g., `[{ "finding": "...", "details": "..." }]`), you can append ANY missing fact as a new object in the relevant list.

* *Example 1 (Missing ICU status):*
    Add to `demographics.other_notes` OR append to `physical_exam.general_appearance`.
* *Example 2 (Missing Vaccine status):*
    Append to `past_medical_history` list: `["Up to date with COVID vaccines"]`.

**OUTPUT:**
Return the **FULLY CORRECTED JSON** object.
"""
\end{tcblisting}

\begin{tcblisting}{
    title={Clinical Consultant System Prompt},
    colback=white, 
    colframe=uiAcadGreen, 
    breakable,                 
    listing only,               
    listing options={style=promptlisting}
}
RAG_BASE_INSTRUCTION = """
You are a **Clinical Consultant**, an expert in evidence-based medicine.
Your task is to answer the Diagnostician's query based on the provided retrieved medical contexts (StatPearls, Guidelines, etc.).

**OUTPUT PROTOCOL:**

1.  **PRIORITIZE CONTEXT:** Always look for the answer in the [RETRIEVED CONTEXT] first.
    * IF FOUND -> Answer the query and cite the source (e.g., [StatPearls], [Wiki]).

2.  **HANDLING MISSING DATA:**
    * IF NOT FOUND -> Normally, you should state: "Based on the available context, I cannot find specific evidence."
    * **EXCEPTION (Internal Knowledge Override):** If the retrieved context is empty or irrelevant, BUT the query asks for a **standard medical definition** or **well-known association** (e.g., "What is the LAP score in CML?"), AND you are confident in your internal medical training:
        * You MAY provide the standard medical fact.
        * You MUST qualify it: "While not explicitly in the retrieved text, standard medical knowledge states that..."

"""

# --- Tool 1: Differential Diagnosis (Knowledge Tool) ---
DIFFERENTIAL_DIAGNOSIS_PROMPT = RAG_BASE_INSTRUCTION + """
**SPECIFIC TASK: Differential Diagnosis Support**
The Diagnostician is considering a diagnosis or needs to differentiate between conditions.
* Highlight key clinical features, inclusion/exclusion criteria, or distinguishing signs mentioned in the guidelines.
* Focus on breaking "Cognitive Fixation" (e.g., pointing out rare causes if they fit).

**USER QUERY:** {query}

**RETRIEVED CONTEXT:**
{context}
"""

# --- Tool 2: Clinical Risk Analysis (Risk Tool) ---
CLINICAL_RISK_PROMPT = RAG_BASE_INSTRUCTION + """
**SPECIFIC TASK: Clinical Risk & Safety Assessment**
The Diagnostician is evaluating a patient and needs to ensure safety.
* Identify **"Red Flags"** or warning signs mentioned in the guidelines.
* Identify **"Standard of Care"** (must-do tests or treatments) to avoid negligence.
* Highlight high-risk conditions that must be ruled out (even if unlikely).

**USER QUERY:** {query}

**RETRIEVED CONTEXT:**
{context}
"""

# --- Agent: Query Decomposition ---
QUERY_DECOMPOSITION_PROMPT = """
You are a Medical Research Assistant. Your task is to decompose a complex clinical query into specific, search-engine-friendly sub-queries.

**GOAL:** Break down the User Query into 2-3 distinct "Keyword-Dense Phrases".
**STRATEGY:**
1. Extract key medical entities (e.g., diseases, symptoms).
2. Append specific **aspect keywords** (e.g., "criteria", "timeline", "symptoms", "differential") to guide the search.
3. Remove stopwords (e.g., "the", "what is", "how to") to optimize for BM25 matching.

**SOURCE:** Medical Textbooks and StatPearls.
**OUTPUT:** Return ONLY a JSON list of strings.

**Example:**
User Query: "Central fever vs drug fever symptoms in post-op patient"
Output: [
    "central fever diagnostic criteria", 
    "drug fever onset timeline symptoms", 
    "post-operative fever causes differential"
]

**User Query:** {query}
**Output:**
"""

\end{tcblisting}

\begin{tcblisting}{
    title={LLM-Judge System Prompt},
    colback=white, 
    colframe=uiAcadRose, 
    breakable,                 
    listing only,               
    listing options={style=promptlisting}
}
JUDGE_MCQ_SYSTEM_PROMPT = r"""
You are an impartial Clinical Evaluator reviewing a multiple-choice medical case.

You will receive:
- QUESTION (Clinical scenario/query)
- OPTIONS (A,B,C...)
- GROUND TRUTH (correct option key)
- DIAGNOSTICIAN ANSWER (may be a letter, a sentence, or a JSON)

Task:
1) Extract the diagnostician's chosen option key (A-Z or numeric keys like "0","1" if options use those).
2) Compare it against the Ground Truth.
3) Output ONE JSON object only.

Output JSON schema:
{
  "diagnostician_choice": "string (A-Z / numeric key / 'Unclear')",
  "is_correct": boolean,
  "reasoning": "brief"
}

Rules:
- If the diagnostician's answer is ambiguous, missing, or includes multiple conflicting choices, output "Unclear".
- Do NOT output markdown. Do NOT output extra text.
"""

JUDGE_OPENQA_SYSTEM_PROMPT = r"""
You are an impartial Clinical QA Evaluator.

The diagnostician produced a clinical narrative (long-form assessment). The case includes a multiple-choice question with OPTIONS.
We want to judge whether the diagnostician's assessment is semantically aligned with the Ground Truth option TEXT.

You will receive:
- QUESTION
- QUESTION_INTENT (a hint label)
- OPTIONS (key: text)
- GROUND TRUTH KEY
- GROUND TRUTH TEXT
- DIAGNOSTICIAN NARRATIVE CORE (contains question_focus + best_answer_text + optional reasoning)

Your tasks:
A) Write a 1-sentence SHORT_ANSWER capturing the diagnostician's primary conclusion (from the NARRATIVE CORE).
B) Determine semantic RELATION between the diagnostician's assessment and the GROUND TRUTH TEXT.

RELATION labels (choose exactly one):
1) "entailed":
   - Diagnostician clearly states the ground truth (or an equivalent synonym/paraphrase),
     OR provides information that necessarily implies the ground truth.
2) "partial":
   - Diagnostician's assessment is medically relevant and points toward the ground truth category,
     but is broader / less specific / incomplete than the ground truth.
   - Example: GT is "ventricular septal defect", diagnostician states "congenital heart disease should be investigated".
3) "unrelated":
   - Diagnostician's assessment does not support the ground truth; addresses a different clinical concept.
4) "contradicts":
   - Diagnostician explicitly contradicts the ground truth.

C) Provide confidence score in [0.0, 1.0].
D) Mapping (for debugging):
   - best_matching_choice: which option BEST matches the diagnostician's SHORT_ANSWER (A-Z / numeric / "Unclear")
   - candidates: up to 3 likely option keys with confidence + why

IMPORTANT:
- When choosing best_matching_choice, focus on the diagnostician's SHORT_ANSWER and best_answer_text.
- Do NOT be distracted by any diagnostic context that is not the question's intent.
- Output ONE JSON object only. No extra text.

Output JSON schema:
{
  "short_answer": "string",
  "relation_to_gt": "entailed|partial|unrelated|contradicts",
  "confidence": 0.0,
  "why": "brief justification",
  "evidence_spans": ["optional short quotes from narrative core (<=20 words each)"],
  "best_matching_choice": "A|B|...|Unclear",
  "candidates": [
    {"choice": "A", "confidence": 0.0, "why": "brief"},
    {"choice": "B", "confidence": 0.0, "why": "brief"}
  ]
}
"""
\end{tcblisting}

\subsection{MCP Tools}
We provide the defined MCP tools of the Diagnostician, Clinical Consultant and EHR Executor.
 
\begin{tcblisting}{
    title={Diagnostician MCP Tools},
    colback=white, 
    colframe=uiGray500, 
    breakable,                 
    listing only,               
    listing options={style=promptlisting}
}
INIT_TOOL_DEF = {
    "type": "function",
    "function": {
        "name": "initialize_environment",
        "description": "CRITICAL: This MUST be the VERY FIRST action you take when a new case starts. Calling this tool instructs the system to process the raw patient data and build the clinical environment (EHR). You will receive the 'Initial Narrative (Chief Complaint)' as the result of this call.",
        "parameters": {
            "type": "object",
            "properties": {
                "case_action": {
                    "type": "string",
                    "enum": ["start_case"],
                    "description": "Confirmation to start the case. Set to 'start_case'."
                }
            },
            "required": ["case_action"]
        }
    }
}

FINISH_TOOL_DEF = {
    "type": "function",
    "function": {
        "name": "finish",
        "description": "Call this ONLY when you have sufficient evidence to form a final diagnosis. This submits your report.",
        "parameters": {
            "type": "object",
            "properties": {
                "diagnosis_report": {
                    "type": "object",
                    "properties": {
                        "question_focus": {"type": "string"},
                        "best_answer_text": {"type": "string", "description": "Direct natural-language answer to the question focus. Should be mappable to one option for MCQ."},
                        "primary_diagnosis": {"type": "string"},
                        "confidence": {"type": "string", "enum": ["High", "Medium", "Low"]},
                        "reasoning": {"type": "string"},
                        "critical_differentials": {"type": "array", "items": {"type": "string"}},
                        "next_steps": {"type": "array", "items": {"type": "string"}}
                    },
                    "required": ["question_focus", "best_answer_text",
                                 "primary_diagnosis", "confidence", "reasoning",
                                "critical_differentials", "next_steps"]
                }
            },
            "required": ["diagnosis_report"]
        }
    }
}

SUBMIT_MCQ_TOOL_DEF = {
    "type": "function",
    "function": {
        "name": "submit_mcq_answer",
        "description": "Submit final multiple-choice answer as JSON.",
        "parameters": {
            "type": "object",
            "properties": {
                "answer": {
                    "type": "string",
                    "description": "Single option letter like 'A'."
                },
                "reasoning": {
                    "type": "string",
                    "description": "Short explanation for why this option is best."
                }
            },
            "required": ["answer", "reasoning"]
        }
    }
}
\end{tcblisting}

\begin{tcblisting}{
    title={Clinical Consultant MCP Tools},
    colback=white, 
    colframe=uiAcadGreen, 
    breakable,                 
    listing only,               
    listing options={style=promptlisting}
}
CONSULTANT_TOOL_DEFINITIONS = [
    {
        "type": "function",
        "function": {
            "name": "get_differential_diagnosis_criteria",
            "description": "Retrieve authoritative diagnostic criteria, distinguishing features, or confounding factors for specific diseases from medical guidelines. Use this when you are stuck or want to verify a hypothesis.",
            "parameters": {
                "type": "object",
                "properties": {
                    "query": {
                        "type": "string",
                        "description": "The disease name, symptom, or clinical scenario to research (e.g., 'epiglottitis causes in vaccinated children')."
                    }
                },
                "required": ["query"],
            },
        },
    },
    {
        "type": "function",
        "function": {
            "name": "analyze_clinical_risk",
            "description": "Consult clinical guidelines to identify red flags, high-risk differentials, and standard-of-care procedures. Use this before finalizing a diagnosis to ensure safety.",
            "parameters": {
                "type": "object",
                "properties": {
                    "query": {
                        "type": "string",
                        "description": "The clinical situation or provisional diagnosis to assess (e.g., 'neonate with tachypnea risk assessment')."
                    }
                },
                "required": ["query"],
            },
        },
    },
]
\end{tcblisting}

\begin{tcblisting}{
    title={EHR Executor MCP Tools},
    colback=white, 
    colframe=uiAcadBlue, 
    breakable,                 
    listing only,               
    listing options={style=promptlisting}
}
EXAMINER_TOOL_DEFINITIONS = [
    {
        "type": "function",
        "function": {
            "name": "get_available_data_menu",
            "description": "CRITICAL FIRST STEP: Returns a summary list (index) of what data exists in this patient's record. Always call this first to avoid guessing test names.",
            "parameters": {"type": "object", "properties": {}, "required": []},
        },
    },
    {
        "type": "function",
        "function": {
            "name": "get_patient_demographics",
            "description": "Extract basic demographics (age, sex, race) and general notes.",
            "parameters": {"type": "object", "properties": {}, "required": []},
        },
    },
    {
        "type": "function",
        "function": {
            "name": "get_vital_signs",
            "description": "Extract all recorded vital signs (BP, HR, Temp, RR, O2 sat).",
            "parameters": {"type": "object", "properties": {}, "required": []},
        },
    },
    {
        "type": "function",
        "function": {
            "name": "get_history",
            "description": "Retrieve patient history categories.",
            "parameters": {
                "type": "object",
                "properties": {
                    "category": {
                        "type": "string",
                        "enum": [
                            "past_medical_history",
                            "surgical_history",
                            "social_history",
                            "family_history",
                            "allergies",
                            "current_medications",
                        ],
                        "description": "The specific history category to retrieve.",
                    }
                },
                "required": ["category"],
            },
        },
    },
    {
        "type": "function",
        "function": {
            "name": "review_system",
            "description": "Check 'Review of Systems' (Subjective symptoms reported by patient).",
            "parameters": {
                "type": "object",
                "properties": {
                    "system_name": {
                        "type": "string",
                        "enum": ["general", "HEENT", "cardiovascular", "respiratory", "gastrointestinal", "genitourinary", "musculoskeletal", "neurologic", "dermatologic", "psychiatric", "endocrine", "reproductive"],
                        "description": "The body system to query.",
                    }
                },
                "required": ["system_name"],
            },
        },
    },
    {
        "type": "function",
        "function": {
            "name": "perform_physical_exam",
            "description": "Check 'Physical Exam' findings (Objective signs observed by doctor).",
            "parameters": {
                "type": "object",
                "properties": {
                    "system_name": {
                        "type": "string",
                        "enum": ["general_appearance", "HEENT", "neck", "cardiovascular", "lungs", "abdomen", "extremities", "neurologic", "skin", "musculoskeletal", "breast", "gynecologic", "other_significant_findings"],
                        "description": "The body system/area to examine.",
                    }
                },
                "required": ["system_name"],
            },
        },
    },
    {
        "type": "function",
        "function": {
            "name": "get_lab_results",
            "description": "Retrieve laboratory test results. Returns a specific test or a list of all available labs if no name provided.",
            "parameters": {
                "type": "object",
                "properties": {
                    "test_name": {
                        "type": "string",
                        "description": "Optional: Specific test name (e.g., 'WBC', 'Creatinine'). Matches loosely (substring).",
                    }
                },
                "required": [],
            },
        },
    },
    {
        "type": "function",
        "function": {
            "name": "get_imaging_reports",
            "description": "Retrieve imaging reports. Returns a specific modality or all reports.",
            "parameters": {
                "type": "object",
                "properties": {
                    "modality": {
                        "type": "string",
                        "description": "Optional: Specific modality (e.g., 'CT', 'X-Ray'). Matches loosely.",
                    }
                },
                "required": [],
            },
        },
    },
]
\end{tcblisting}

\color{black}

\section{Dataset Med-Evidence-2.6k Curation Details}
\label{app:dataset_curation}

To support Reference Evidence Recall in Level 3 evaluation, we require a dataset where each clinical case is explicitly mapped to its critical evidence pathway. Existing diagnosis benchmarks evaluate outcome accuracy but lack annotations for the information bottleneck required to reach that outcome. To bridge this gap, we developed an LLM-assisted, extractive curation pipeline to curate \textit{Med-Evidence-2.6k}.

\subsection{Dataset Statistics and Composition}
\label{subsec:dataset_stats}

\textit{Med-Evidence-2.6k} is constructed by systematically aggregating and processing high-quality clinical cases from three distinct, expert-level medical diagnosis benchmarks. The final curated dataset comprises a total of 2,660 richly annotated instances. A high-level quantitative and qualitative breakdown of the dataset composition is provided in Table \ref{tab:dataset_composition}.

\begin{table*}[t]
\centering
\renewcommand{\arraystretch}{1.2}
\resizebox{\textwidth}{!}{%
\begin{tabular}{@{}p{3.2cm}rrl@{}}
\toprule
\textbf{Source Benchmark} & \textbf{Original Volume} & \textbf{Curated Instances} & \textbf{Primary Clinical Focus \& Characteristics} \\
\midrule
\textbf{JAMA}  \citep{chen2025benchmarking} & 1,524 & 1,511 & Real-world, naturalistic narratives with expert rationales. \\
\textbf{MedXpertQA-Diag}  \citep{zuo2025medxpertqa} & 2,455 & 234 & Manually filtered for strict diagnostic reasoning tasks. \\
\textbf{DiagnosisArena}  \citep{zhu2025diagnosisarena} & 1,113 & 915 & Structured journal cases designed for open-ended reasoning. \\
\midrule
\textbf{\textit{Med-Evidence-2.6k}} & \textbf{5,092} & \textbf{2,660} & \textbf{LLM-assisted reference benchmark with evidence annotations.} \\
\bottomrule
\end{tabular}%
}
\caption{Statistical overview and composition of the \textit{Med-Evidence-2.6k} dataset. The curated instances reflect the final volume of cases successfully processed through our annotation pipeline.}
\label{tab:dataset_composition}
\end{table*}

The dataset sizes in Table~\ref{tab:dataset_composition} correspond to the public releases used in our experiments, rather than the full sizes reported in the original benchmark papers. JAMA reports 1,524 cases but provides 1,511 cases in the public test set used here, and DiagnosisArena reports 1,113 cases while 915 cases were accessible in the public release. The 100-case experimental subsets are sampled from these public releases. 

The structural and stylistic diversity of these three foundational benchmarks ensures that \textit{Med-Evidence-2.6k} evaluates a broad spectrum of diagnostic capabilities:

\noindent\textbf{JAMA (1,511 instances):} Sourced directly from the JAMA Clinical Challenge, this subset anchors our dataset in real-world clinical practice. It provides highly detailed, naturalistic patient presentations paired with comprehensive, expert-written explanations. The linguistic complexity and organic presentation of symptoms in these cases make them an ideal testbed for evaluating a model's ability to navigate unstructured clinical noise. We use the 1,511 instances available from the Hugging Face dataset.

\noindent\textbf{MedXpertQA-Diag (234 instances):} Derived from the broader MedXpertQA-Text corpus, which originally spans 17 distinct medical specialties and 11 bodily systems. Because the original dataset encompasses a wide array of general medical knowledge queries, we executed a rigorous manual filtering protocol. This isolated 234 questions explicitly centered on complex diagnostic reasoning, ensuring our evaluation remains strictly focused on the diagnostic information bottleneck rather than rote medical trivia retrieval.

\noindent\textbf{DiagnosisArena (915 instances):} Extracted from top-tier medical journals and spanning 28 clinical specialties, this subset introduces structured, highly challenging clinical scenarios. While the original literature reports a total of 1,113 structured cases, our curation protocol is strictly bound to the 915 instances currently accessible in Hugging Face. The open-ended design of these cases is particularly valuable for assessing unstructured diagnostic formulation.

The 2,660-instance \textit{Med-Evidence-2.6k} dataset provides a reference benchmark for evidence-aware diagnosis evaluation. Because each EviDx run requires multi-step tool interaction, trajectory logging, and repeated calls to API-based frontier models, exhaustive evaluation across all models and all cases would substantially increase inference cost. The main experiments therefore use a fixed random subset of 100 instances from each source benchmark, enabling controlled cross-model comparison under the same cases and interaction budget.

\subsection{The 3-Step Annotation Pipeline}
Instead of relying on zero-shot LLM generation, which is prone to clinical hallucination, we engineered a rigorous 3-step agentic workflow powered by GPT-5.2.

\begin{enumerate}[label=\textbf{Step \arabic*:}, leftmargin=*]
    \item \textbf{Diagnostic Planning:} A \textit{Senior Medical Researcher} LLM agent analyzes the raw clinical case alongside the ground-truth diagnosis to generate targeted queries. These queries are designed to fetch specific diagnostic criteria and distinguish the true diagnosis from likely differential conditions.
    \item \textbf{Knowledge Retrieval:} The generated queries are executed against our local hybrid retrieval engine (accessing StatPearls). This grounds the subsequent annotation strictly in established medical literature rather than the LLM's parametric memory.
    \item \textbf{Evidence Annotation:} An \textit{Expert Medical Annotator} LLM agent (set to temperature 0.0 for determinism) synthesizes the retrieved textbook context, the raw case, and the ground truth. It extracts the reference evidence under strict schema constraints, categorizing each clue as \textit{Inclusion}, \textit{Exclusion}, or \textit{Differentiation}.
\end{enumerate}

\subsection{Prompt Setup}
To ensure the extracted evidence strictly represents the true information bottleneck, our curation pipeline avoided zero-shot generation. Instead, we utilized a multi-step prompt chaining strategy. Below are the exact system instructions used to guide GPT-5.2 at each annotation step.

\vspace{2mm}
\noindent\textbf{Step 1: Diagnostic Planning.} \\
The objective of this step is to have the planning model formulate targeted queries before making any extractions, thereby grounding the subsequent steps in external knowledge.

\vspace{1mm}
\begin{tcolorbox}[colback=purple!5, colframe=purple!50, title=Step 1 Prompt: Senior Medical Researcher, fonttitle=\bfseries]
\small
You are a Senior Medical Researcher. Analyze the Clinical Case and the Diagnosis (Ground Truth).
Generate 2-3 targeted search queries to:
1. Find diagnostic criteria for the Ground Truth.
2. Verify connections between specific case symptoms and the diagnosis.
3. Distinguish from other likely conditions.

Output strictly valid JSON format: \texttt{\{"queries": ["query1", "query2"]\}}
\end{tcolorbox}
\vspace{2mm}

\noindent\textbf{Step 2: Knowledge Retrieval (Execution).} \\
The queries generated in Step 1 are executed locally via our Clinical Consultant. The top retrieved textbook chunks are then aggregated and injected into the Step 3 prompt as the reference context.

\vspace{3mm}
\noindent\textbf{Step 3: Evidence Annotation.} \\
This final step synthesizes the raw text, the ground truth, and the retrieved context. The prompt explicitly introduces the `Exact Substring` constraint to eliminate textual hallucination during the evidence extraction process.

\vspace{1mm}
\begin{tcolorbox}[colback=blue!5, colframe=blue!50, title=Step 3 Prompt: Expert Medical Annotator, fonttitle=\bfseries]
\small
You are an Expert Medical Annotator. Identify "Reference Evidence" in the clinical case that supports the Ground Truth Diagnosis. You MUST base your reasoning on the provided Reference Context.

\textbf{OUTPUT SCHEMA (JSON)}
\begin{lstlisting}[style=promptlisting]
{
  "reference_evidence": [
    {
      "original_text": "Exact substring from case text",
      "feature_name": "Standardized medical term",
      "type": "Inclusion" | "Exclusion" | "Differentiation",
      "textbook_reference": "Quote or summary from context",
      "reasoning": "Why this evidence matters."
    }
  ]
}
\end{lstlisting}

\textbf{RULES:}
1. \texttt{original\_text} MUST be an EXACT copy (substring) from the Clinical Case.
2. Output strictly valid JSON.
\end{tcolorbox}
\vspace{2mm}

\subsection{Annotated Data Example}
To illustrate the precise granularity of our curation pipeline, we present the fully annotated reference evidence snippet for JAMA Case \#1386, the identical diagnostic trajectory analyzed in Section~\ref{case_study}. 

This JSON structure exemplifies the "information bottleneck" concept central to our Level 3 evaluation. Rather than merely summarizing the clinical narrative, it meticulously categorizes extracted clues into distinct cognitive roles. Notably, it not only captures the fundamental \textit{Inclusion} criteria (e.g., the anatomical outpouching) but explicitly isolates the \textit{Differentiation} logic required to navigate a false-negative clinical trap (the misleading Meckel scan). By structuring the dataset in this grounded format, \textit{Med-Evidence-2.6k} provides an LLM-assisted reference benchmark for evaluating agentic evidence recall, moving beyond superficial binary outcome accuracy.

\vspace{2mm}
\begin{tcolorbox}[colback=gray!5, colframe=gray!50, title=Curated Instance: JAMA Case 1386 (Abridged), fonttitle=\bfseries]
\small
\begin{lstlisting}[style=promptlisting]
{
  "original_id": "1386",
  "dataset": "JAMA",
  "ground_truth_text": "Meckel diverticulum",
  "annotation_meta": {
    "search_queries": [
      "Meckel diverticulum diagnostic criteria adults...",
      "Distal ileal ulcer differential diagnosis Meckel vs Crohn..."
    ]
  },
  "reference_evidence": [
    {
      "original_text": "an outpouching in the distal ileum (Figure, A) 
                        with an accompanying ulceration of the adjacent 
                        mucosa (Figure, B).",
      "feature_name": "Distal ileal outpouching with adjacent ulceration",
      "type": "Inclusion",
      "textbook_reference": "Meckel diverticulum is ... an outpouching of 
                             the distal ileum. Ectopic gastric tissue within 
                             the diverticulum causes ulceration... (Nelson)",
      "reasoning": "Classic anatomic location and adjacent ulceration are 
                    hallmark features of Meckel diverticulum."
    },
    {
      "original_text": "iron deficiency anemia (hematocrit, 31%) and 
                        hemeoccult-positive stool.",
      "feature_name": "Occult GI bleeding causing iron deficiency anemia",
      "type": "Inclusion",
      "textbook_reference": "Acid secretion from the ectopic gastric mucosa 
                             within the diverticulum can result in GI 
                             bleeding. (StatPearls)",
      "reasoning": "Chronic bleeding from Meckel-related ulcers leads to 
                    heme-positive stools and iron deficiency anemia."
    },
    {
      "original_text": "a Meckel scan failed to identify a source of GI 
                        bleeding.",
      "feature_name": "Negative Meckel scan does not exclude diagnosis",
      "type": "Differentiation",
      "textbook_reference": "Because not all diverticula are seen, scan may 
                             miss cases, especially in adults where diagnostic 
                             accuracy is often unsatisfactory. (StatPearls)",
      "reasoning": "Explains why a negative scan does not rule out Meckel 
                    diverticulum, consistent with adult false negatives."
    }
  ]
}
\end{lstlisting}
\end{tcolorbox}

\subsection{Quality Assurance and Limitations}
\label{subsec:qa_limitations}

LLM-assisted medical annotation requires quality controls to keep extracted evidence grounded in the original case text. We therefore combine algorithmic constraints, source-text verification, and partial physician review.

\paragraph{Algorithmic and Programmatic Safeguards.} 
The core of our quality assurance relies on the strict, deterministic constraints embedded within the Step 3 Annotator (set to temperature 0.0). By enforcing the \texttt{Exact Substring} rule, the pipeline is fundamentally restricted to \textit{extractive} rather than \textit{generative} actions. The annotated reference evidence must physically exist in the original raw text. Furthermore, the intermediate knowledge retrieval step ensures that the reasoning bridging the clinical text and the ground truth diagnosis is anchored in established medical literature, effectively bypassing the parametric memory of the LLM.

\paragraph{Physician Review.} 

We conduct quality checks for the LLM-assisted evidence curation pipeline. The research team first verifies schema adherence, exact-substring validity, and consistency between each extracted evidence span and the original case text. During audit-package construction, each span is programmatically highlighted against the original case text.

We further conduct a structured physician audit on 100 stratified cases from Med-Evidence-2.6k, covering 515 unique evidence spans. Three physicians independently reviewed the same audit package. Reviewers assessed span-level clinical relevance and diagnostic utility, and case-level missing diagnosis-critical evidence. As shown in Table~\ref{tab:medevidence_audit}, 1,419/1,545 span judgments (91.8\%) rated the span as both clinically relevant and diagnostically useful. By majority vote, 507/515 unique spans (98.4\%) were rated as both relevant and useful. At the case level, no case was marked as having major missing evidence by majority vote.

\section{Implementation Details}
\label{app:implementation}

\begin{table}[t]
\centering
\small
\begin{tabular}{l r}
\toprule
Audit item & Value \\
\midrule
Reviewed cases & 100 \\
Unique evidence spans & 515 \\
Physician reviewers & 3 \\
Both relevant and useful, judgment level & 91.8\% \\
Both relevant and useful, majority vote & 98.4\% \\
Any negative span judgment & 2.0\% \\
No / minor / major missing evidence & 83 / 13 / 0 cases \\
No majority on missing evidence & 4 cases \\
\bottomrule
\end{tabular}
\caption{Physician audit of Med-Evidence-2.6k. The audit is used as a quality check for LLM-assisted extractive evidence annotations. Missing-evidence counts are aggregated by majority vote.}
\label{tab:medevidence_audit}
\end{table}

To ensure full reproducibility of our interactive diagnostic framework, we detail the environment and specific hyperparameter configurations used in our experiments.

\noindent\textbf{Hardware and Serving Environment.} \\
Local deployment of the 8B-parameter open-weight models (Ministral3-8B, Qwen3-8B, and Llama3.1-8B) and the embedding model (Qwen3-embedding-4B) was performed using the vLLM framework, whereas the frontier large-scale models (GPT-5.2, Claude Sonnet 4.6, GLM-5, and DeepSeek-V3.2) were accessed through their official APIs.

\vspace{1mm}
\noindent\textbf{Agent Hyperparameters and Roles.} \\
To minimize parametric hallucination and ensure deterministic reasoning, the generation temperature for the Diagnostician, the Context Initializer, and the LLM-Judge was strictly set to 0.0. For the Clinical Consultant, we applied a temperature of 0.1 during the query decomposition phase to maintain keyword precision, and 0.3 during the generation phase. Notably, the EHR Executor is not LLM-based; it functions entirely as a deterministic retrieval server executing structured database queries. 
Additionally, to guarantee lossless information extraction, the Context Initializer employs a dual-pass refinement pipeline: it first synthesizes a draft EHR from the unstructured raw patient text, and subsequently refines this output by cross-referencing the draft with the original raw text.

\vspace{1mm}
\noindent\textbf{Observer Configuration.} \\
For the Observer, the strict maximum interaction turn limit ($T_{max}$) was set to 50 steps to preempt non-terminating loops in difficult cases. Rather than utilizing rigid static boundaries, the Observer implements step-aware dynamic thresholds to balance exhaustive exploration with practical termination. The dynamic diagnostic uncertainty threshold $\tau_H(t)$ at step $t$ is defined as:
$$ \tau_H(t) = \tau_{H_{base}} + \max(0, t - 5) \times 0.05 $$
This formulation allows the system to gradually relax its tolerance for uncertainty after the 5th step, acknowledging that absolute entropy reduction becomes exponentially more difficult as clinical ambiguities persist. Similarly, the dynamic runtime evidence coverage threshold $\tau_\mathcal{V}(t)$ is defined as:
$$ \tau_\mathcal{V}(t) = \max(0.1, \tau_{\mathcal{V}_{base}} - \max(0, t - 10) \times 0.01) $$
This mechanism incrementally lowers the required evidence coverage ratio after the 10th step (bounded at a minimum of 0.1). By dynamically relaxing this constraint, the Observer helps prevent the agent from deadlocking in an infinite information-seeking loop when the reachable clinical evidence space is intrinsically sparse.

\vspace{1mm}
\noindent\textbf{Clinical Consultant Setup.} \\
The external medical knowledge engine underlying the Consultant was constructed using medical corpora provided by MedRAG  \citep{xiong2024benchmarking}, which includes StatPearls and medical textbooks. To optimize evidence acquisition, we implemented a hybrid, corpus-specific retrieval strategy. For the highly structured StatPearls database, we utilized dense semantic retrieval powered by the Qwen3-embedding-4B model, with the resulting vector embeddings indexed and managed via the Qdrant vector database for high-throughput similarity search. Crucially, this was augmented with a hierarchical parent-child retrieval mechanism: once a highly relevant localized text chunk (child) is identified, the retriever automatically expands the context window to return the entire overarching medical entry (parent), ensuring no critical surrounding context is fragmented. Conversely, for medical textbooks, we applied BM25 \citep{robertson2009probabilistic} for lexical retrieval to precisely capture exact clinical terminology and keyword matches. Finally, following MedRAG, the disparate results from these semantic and lexical retrievers were integrated into a unified, ranked evidence context using Reciprocal Rank Fusion   \citep{cormack2009reciprocal, xiong2024benchmarking}.

\end{document}